\documentclass{article}
\usepackage{ijcai23}

\usepackage{times}
\usepackage{soul}
\usepackage{url}
\usepackage[hidelinks]{hyperref}
\usepackage[utf8]{inputenc}
\usepackage[small]{caption}
\usepackage{graphicx}
\usepackage{amsmath}
\usepackage{amsthm}
\usepackage{booktabs}
\usepackage{algorithm}
\usepackage{algorithmic}
\usepackage[switch]{lineno}
\usepackage{subfigure}
\usepackage{adjustbox}
\usepackage{color}
\title{Large Language Model Guided Knowledge Distillation for Time Series \\ Anomaly Detection}

\author{
    Chen Liu, Shibo He, Qihang Zhou, Shizhong Li, Wenchao Meng
    \affiliations
    Zhejiang University
    \emails
    \{liu777ch, s18he, zqhang, lisz, wmengzju\}@zju.edu.cn
}

\begin{document}

\maketitle

\begin{abstract}
Self-supervised methods have gained prominence in time series anomaly detection due to the scarcity of available annotations.
Nevertheless, they typically demand extensive training data to acquire a generalizable representation map, which conflicts with scenarios of a few available samples, thereby limiting their performance.
To overcome the limitation, we propose \textbf{AnomalyLLM}, a knowledge distillation-based time series anomaly detection approach where the student network
is trained to mimic the features of the large language model (LLM)-based teacher network that is pretrained on large-scale datasets.
During the testing phase, anomalies are detected when the discrepancy between the features of the teacher and student networks is large. 
To circumvent the student network from learning the teacher network's feature of anomalous samples, we devise two key strategies. 1) Prototypical signals are incorporated into the student network to consolidate the normal feature extraction. 2) We use synthetic anomalies to enlarge the representation gap between the two networks. AnomalyLLM demonstrates state-of-the-art performance on 15 datasets, improving accuracy by at least 14.5\% in the UCR dataset.

\end{abstract}

\section{Introduction}

Time series anomaly detection (TSAD) aims to identify abnormal data whose patterns deviate from the majority of the data \cite{blazquez2021review}. It plays critical roles in numerous applications such as industrial fault diagnosis, network intrusion detection, and health monitoring \cite{kieu2022anomaly}. 

The primary challenge for TSAD lies in the laborious process of acquiring annotations \cite{ruff2021unifying}. Consequently, most previous works follow the unsupervised setting where no labels are provided, and the majority of the data is assumed to be normal \cite{audibert2020usad}. These methods can be broadly categorized into one-class classification-based methods \cite{ruff2018deep,shen2020timeseries,carmona2021neural}, density estimation-based methods \cite{dai2022graph,zhou2023detecting}, and self-supervised methods \cite{jeong2023anomalybert}. With the advancement of representation learning, self-supervised methods have garnered growing attention and dominated the field \cite{zhang2023self}. They employ pretext tasks such as reconstruction \cite{su2019robust,xu2021anomaly,song2023memto}, forecasting \cite{deng2021graph,li2023staged}, imputation \cite{chen2023imdiffusion}, and contrastive learning \cite{yang2023dcdetector,wang2023deep} to learn a representation map that distinguishes the normal and abnormal samples. 

\begin{figure}[t]
\centering
\includegraphics[width=0.5\textwidth]{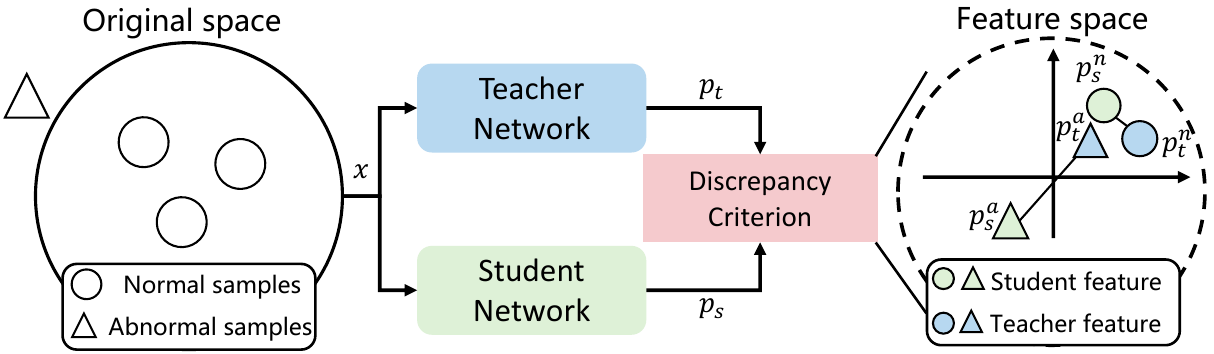}
\caption{Knowledge distillation-based framework: the discrepancy between outputs of the student and teacher networks is expected to be small on normal samples while large on abnormal samples.}\label{fig:idea}
\vspace{-1em}
\end{figure}

However, learning generalizable representations typically requires a vast amount of training data \cite{zhang2023self}, which conflicts with scenarios of limited available samples, thereby limiting the performance of these self-supervised methods. To overcome this limitation, we introduce AnomalyLLM, a novel approach that integrates knowledge distillation and large language models (LLMs). The fundamental idea is to train a student network to mimic the output of a teacher network that is pretrained on a large-scale dataset. During the testing phase, anomalies are identified when a significant discrepancy exists between the outputs of the student and teacher networks, as shown in Fig.~\ref{fig:idea}. To this end, we need to address two key challenges.

\textbf{\emph{How to pretrain the teacher network without large-scale time series datasets?}}
Large-scale datasets for pretraining are abundant in computer vision (CV) and natural language processing (NLP), playing a critical role in generalizable representation learning. However, there is currently a lack of universal large-scale time series datasets, with the largest available dataset being less than 10GB, a size significantly smaller than that in CV and NLP \cite{godahewa2021monash}. Consequently, pretraining the teacher network remains a challenge. Recent studies have explored the modal similarity between language and time series, revealing the remarkable potential of pretrained LLMs in generating time series representations \cite{zhou2023one}. LLMs can be fine-tuned on time series data under few-shot \cite{jin2023time} or even zero-shot \cite{gruver2023large} settings. This observation motivates us to use a pretrained LLM as the teacher network. We follow the time series embedding layer with a pretrained LLM, adapting it to generate time series representations.

\textbf{\emph{How to circumvent the student network from 'overlearning' representations produced by the teacher network?}}
We anticipate the discrepancy between the outputs of the student and teacher networks to be small on normal samples but large on abnormal samples \cite{zhou2022pull}. However, given the absence of abnormal samples to enlarge their representation gap, this can easily lead to overlearning of the student network, where the two networks consistently generate similar representations, even for abnormal samples. To circumvent this problem, we implement two designs. First, we incorporate prototypical signals into the student network, enabling its representations to focus more on the historical normal patterns \cite{song2023memto}. Second, we employ data augmentations to produce synthetic anomalies \cite{sun2023unraveling}, which are used to enlarge the representation discrepancy. Furthermore, the teacher network’s representations of original and augmented samples are treated as positive pairs, and a contrastive loss is applied to bring them closer together, serving as a regularization term to encourage the teacher network to capture more general patterns.
Comprehensive experiments are conducted to demonstrate the superiority of our method on 9 univariate datasets and 6 multivariate datasets.

The main contributions are summarized as follows:

\begin{itemize}
\item As far as we know, AnomalyLLM is the first knowledge distillation-based time series anomaly detection method.  

\item We devise a teacher network that is adapted from the pretrained LLM, capable of learning a rich generalizable representation for time series after fine-tuning.


\item To maintain the discrepancy between the teacher and student networks, we integrate prototypical signals into the student network and design a data augmentation-based training strategy.

\item Extensive experiments show that the proposed model achieves SOTA performance on 15 real-world datasets.

\end{itemize}

\section{Related works}
 
\subsection{Time Series Anomaly Detection}
Time series anomaly detection plays a pivotal role in various real-world applications \cite{kieu2022anomaly}. Early studies employ statistical methods or machine learning-based methods \cite{blazquez2021review}, which fail to describe complex patterns of time series signals. In recent years, with the success of neural networks such as variational autoencoder \cite{park2018multimodal} and generative adversarial network \cite{zhou2019beatgan}, numerous deep learning-based methods have emerged for time series anomaly detection. These methods can be roughly categorized into one-class classification-based methods \cite{ruff2018deep,shen2020timeseries,carmona2021neural}, density estimation-based methods \cite{dai2022graph,zhou2023detecting,zhou2024label}, and self-supervised-based methods \cite{jeong2023anomalybert}. 
With the rapid development of representation learning, self-supervised methods have dominated the field. Reconstruction is the most usual self-supervised method, where the reconstruction error indicates the outlyingness of the samples \cite{xu2021anomaly,li2023prototype,song2023memto}. Forecasting \cite{deng2021graph}, imputation \cite{chen2023imdiffusion}, and contrastive learning \cite{yang2023dcdetector,wang2023deep,sun2023unraveling} also emerge as other pretext tasks for self-supervised anomaly detection. While they have achieved SOTA results on various datasets, the small data size used by these self-supervised methods hinders them from learning generalizable representations, thereby limiting their performance \cite{zhang2023self}.
In this paper, we introduce a teacher network adapted from LLM, which has demonstrated a strong ability to generate generalizable time series representations. A student network is trained to mimic the output of the teacher network, and the discrepancy between their outputs serves as the anomaly score in the testing phase.

\subsection{Large Language Model}
Pretrained foundation models have proven excellent performance in NLP and CV, prompting its progress in time series analysis \cite{jin2023large}. 
Despite the increasing interest in foundation models for time series \cite{garza2023timegpt}, it remains a significant challenge due to the limited availability of large-scale datasets. The largest time series dataset is currently less than 10GB, a size smaller than that of NLP datasets \cite{godahewa2021monash}. However, recent studies suggest that pretrained LLMs can be adapted to time series analysis through fine-tuning on time series data \cite{zhou2023one}. Gruver \emph{et~al.} \cite{gruver2023large} even argue that a pretrained LLM can serve as a zero-shot time series forecaster, thanks to its capability to model flexible distributions over sequences of numbers. Consequently, various studies leverage LLM for time series analysis, with a predominant focus on forecasting \cite{jin2023time,cao2023tempo} and classification \cite{sun2023test}. Notably, GPT4TS \cite{zhou2023one} stands as the sole LLM-based time series anomaly detection method, employing an encoder-decoder reconstruction architecture with GPT2 as the encoder. However, the pretrained GPT2's strong generalization ability makes it prone to reconstructing abnormal signals and yielding false negatives. In contrast, we propose a novel knowledge distillation-based method, where the student network is trained from scratch, which will not generalize to those unseen anomalies compared to GPT4TS.

\subsection{Knowledge Distillation}
Knowledge distillation is proposed by \cite{hinton2015distilling}, aiming to push the student network to regress the output of the teacher network.
It is first employed in vision anomaly detection by \cite{bergmann2020uninformed}.
The fundamental principle is that anomalies are identified when there is a substantial discrepancy between the outputs of the student and teacher networks. While this principle has been widely explored in vision anomaly detection, \cite{salehi2021multiresolution,zhou2022pull}, its application in time series remains an unexplored territory. 
Our method can be seen as the first attempt to introduce knowledge distillation into time series anomaly detection. Moreover, our method diverges from existing works in CV in two key aspects. First, unlike in CV where large datasets are commonly available for pretraining the teacher network, large time series datasets are scarce, impeding the effective pretraining. In this work, we employ the pretrained LLM as our teacher network and adapt time series signals to features LLM can understand by an input embedding layer. Second, to prevent the student network from overlearning the representation of the teacher network, we propose reminding the student network of the prototypical signals and devising a data augmentation-based training strategy. Through these efforts, we demonstrate that knowledge distillation can provide another approach for time series anomaly detection, which has not been explored previously.



\begin{figure*}[t]
\centering
\includegraphics[width=0.8\textwidth]{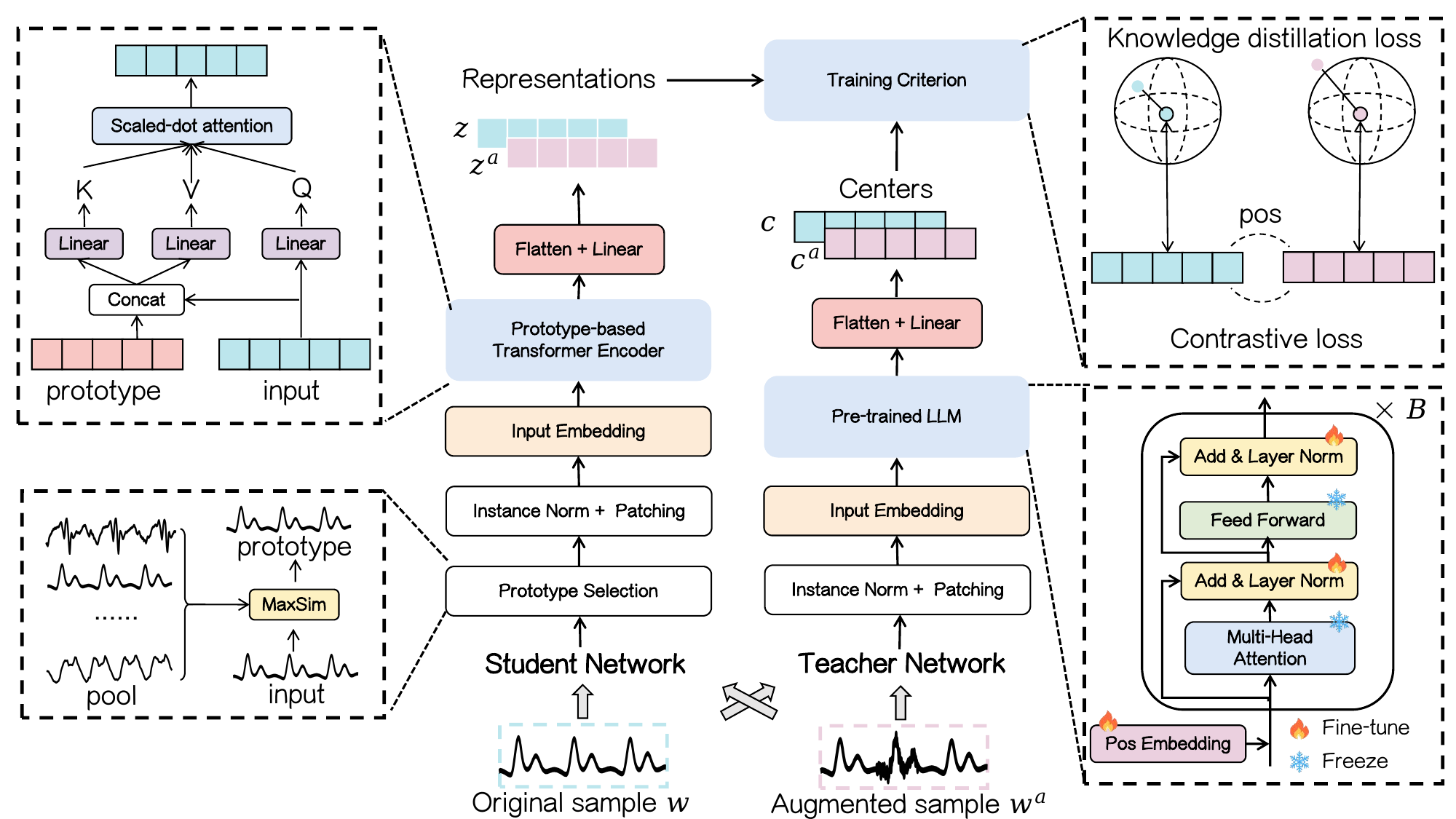}\label{fig:model}
\caption{The framework of AnomalyLLM. It consists of three main components: prototype-based student network, LLM-based teacher network, and data augmentation-based training strategy. }
\vspace{-1em}
\end{figure*}

\section{Methodology}
Given a $D$-dimension multivariate time series $X=[x_1,x_2,...,x_L] \in \mathbf{R}^{D \times L}$ of length $L$, where $x_t \in \mathbf{R}^{D}$ denotes the data collected at the $t$-th time step and $D$ denotes the number of variables, we aim to train an anomaly detector. During the testing phase, we utilize the trained detector to predict an unseen multivariate time series $\hat{X}=[\hat{x_1},\hat{x_2},...,\hat{x_{L'}}]$ with $\hat{Y}=[\hat{y_1},\hat{y_2},...,\hat{y_{L'}}]$, where $\hat{y_l} \in \left\{0,1\right\} $ indicates whether anomalies occur at the $l$-th time step. 

\subsection{Overall Architecture}
Following previous works \cite{carmona2021neural}, we partition the entire time series into several time windows of fixed length $T$. Given a time window $\mathbf{w}_{i} \in \mathbf{R}^{D \times T}$, we aim to identify whether anomalies occur within it. 
Our method adopts the knowledge distillation architecture \cite{bergmann2020uninformed}. The architecture consists of a student network $\phi: \mathbf{R}^{D \times T} \xrightarrow{} \mathbf{R}^{d}$ and a teacher network $\varphi: \mathbf{R}^{D \times T} \xrightarrow{} \mathbf{R}^{d}$, both transforming original signals into $D$-dimensional vectors. The time window is fed into the two networks, and outputs are denoted as $z_{i} = \phi(\mathbf{w}_i)$ and $c_{i} = \varphi(\mathbf{w}_i)$.  We anticipate these two representations to be close for normal samples and distant for abnormal samples. The hypersphere classifier loss  \cite{ruff2020rethinking} is utilized to calculate the discrepancy between two representations:
\begin{equation}
\mathcal{L} = -(1-y_i)\log \ell(z_{i}, c_{i}) - y_i \log(1-\ell(z_{i},c_{i})),\label{eq:1}
\end{equation}where $\ell(z_{i}, c_{i}) = \exp(-{\Vert z_{i}-c_{i} \Vert}_{2}^{2})$, and $y_{i}$ denotes the ground truth label. In the unsupervised setting, all samples are assumed to be normal, and $y_{i}=0$.

\subsection{Prototype-based Student Network}
To prevent the student network from learning overly generalizable representations like the teacher network, we guide it with prototypes. These prototypes represent characteristic segments in the entire time series and are trainable parameters in our method. We select prototypes that closely resemble the input time window to assist in generating the representation.

Prior study \cite{nie2022time} has demonstrated the effectiveness of channel independence in multivariate time series analysis. Therefore, we select the most similar prototype for each channel.
First, we initialize a prototype pool $\mathcal{M}= \left\{ \mathcal{M}_{1}, \mathcal{M}_{2},..., \mathcal{M}_{D} \right\}$, where $\mathcal{M}_{i}=\left\{\mathbf{m}_{i}^{1}, \mathbf{m}_{i}^{2},...,\mathbf{m}_{i}^{M}\right\}$ represents the collection of $M$ prototypes of length $T$ for the $i$-th channel. Given an input time window $\mathbf{w}$, we calculate the similarity between the time windows of each channel and their corresponding prototypes, and then select the most similar prototype:
\begin{equation}
\mathbf{m}_{j}^{s}= \mathop{\max}_{i=1,2,...,M} sim(\mathbf{m}_{j}^{i}, \mathbf{w}^{j}),
\end{equation}
where $\mathbf{w}^{j}$ is the $j$-th channel of input, and $sim(\cdot,\cdot):\mathbf{R}^{T} \times \mathbf{R}^{T}\xrightarrow{}\mathbf{R}^{+} $ represents the function to measure similarity. We use cosine similarity in the paper. The selected prototype for the entire time window is denoted as $\mathbf{m}_s$. 

Instance normalization \cite{kim2021reversible} is employed to mitigate the distribution shift effect and patching \cite{nie2022time} is utilized to extract local semantic information. Both the input time window and prototype are addressed by these two techniques. It is noteworthy that the hyperparameters of the prototype normalization are the same as those of the input. Next, both the input and prototype are fed into an input embedding layer. The input embedding layer consists of linear probing which extracts the in-patch information and a positional embedding which records the position information of the sequences. The layer generates the input embedding $\mathbf{w}_{e}$ and prototype embedding $\mathbf{m}_{e}$. 

To incorporate prototypical features into the original time window, we devise a prototype-based Transformer encoder. The traditional Transformer encoder is stacked by blocks, each consisting of an attention layer, a feed-forward layer, and layer normalization \cite{song2023memto}. In this paper, we provide information about prototypes for each attention layer. Specifically, the queries and keys are produced for both prototypes and inputs. The correlations between the $i$-th patch of the input embedding $\mathbf{w}_{e}^{i}$ and other patches are calculated as follows:
\begin{equation}
\begin{aligned}
s_{i,t}^{w}=\frac{\exp(\langle \mathbf{q}_{w}^{i}, \mathbf{k}_{m}^{t}\rangle)}{\sum_{j=1}^{n}\exp(\langle \mathbf{q}_{w}^{i}, \mathbf{k}_{w}^{j} \rangle) + \sum_{j=1}^{n}\exp(\langle \mathbf{q}_{w}^{i}, \mathbf{k}_{m}^{j} \rangle) }, \\
s_{i,t}^{m}=\frac{\exp(\langle \mathbf{q}_{w}^{i}, \mathbf{k}_{m}^{t}\rangle)}{\sum_{j=1}^{n}\exp(\langle \mathbf{q}_{w}^{i}, \mathbf{k}_{w}^{j} \rangle) + \sum_{j=1}^{n}\exp(\langle \mathbf{q}_{w}^{i}, \mathbf{k}_{m}^{j} \rangle) },
\end{aligned}
\end{equation}
where $\mathbf{q}_{w}, \mathbf{q}_{m}, \mathbf{k}_{w}, \mathbf{k}_{m}$ represents the input patch query, prototype patch query, input patch key and prototype patch key, respectively, and $\langle,\rangle$ represents the inner product.  $n$ denotes the number of patches in the time window. Next, we calculate the representation of the input as follows:
\begin{equation}
\mathbf{o}= \sum_{t=1}^{n} s_{i,t}^{w}\mathbf{v}_{w}^{t} + \sum_{t=1}^{n} s_{i,t}^{m}\mathbf{v}_{m}^{t},
\end{equation}
where $\mathbf{v}_{w}^{t}, \mathbf{v}_{m}^{t}$ represent the value of input and prototype, respectively. The multi-head mechanism is also utilized to capture patterns of different scales.
A flattened layer and a linear layer are subsequently employed to transform the input into the final representations $z$.

\subsection{LLM-based Teacher Network}
The teacher network is expected to produce generalizable representations. Previous works have unveiled the potential of pretrained NLP models such as GPT2 in time series representation generation \cite{zhou2023one}. As a result, we devise the teacher network based on the pretrained LLM.

The input time series undergoes normalization and patching before being fed into the input embedding layer. 
Notably, the input embedding layer consists only of linear embedding, as positional embedding is inherent in the pretrained LLM.  The linear embeddings in the teacher network and the student network utilize different parameters. The linear embedding in the student network is designed to extract correlations within the time series, while that in the teacher network focuses on transforming original time series signals into a representation comprehensible to the language model.

Next, the preprocessed inputs are fed into a network that retains the positional embedding layer and $B$ self-attention blocks of the pretrained LLM. We use GPT2 in this paper. To preserve the knowledge from pretrained LLM, we freeze the attention layer and the feed-forward layer which are crucial components for sequence modeling. The positional embedding layer and the layer normalization are fine-tuned on the input time series, adapting the LLM to understand the time series representations for anomaly detection. The output of the last self-attention block is fed into a flattened and linear layer to generate the eventual representation $c$.

\begin{table*}[t]
\scalebox{0.85}{
\setlength{\tabcolsep}{0.25em}
\begin{tabular}{c|llll|llll|llll|llll|llll}
\hline
\multicolumn{1}{l|}{} & \multicolumn{4}{c|}{ABP}                                                                                       & \multicolumn{4}{c|}{Acceleration}                                                                              & \multicolumn{4}{c|}{Air Temperature}                                                                           & \multicolumn{4}{c|}{ECG}                                                                                       & \multicolumn{4}{c}{EPG}                                                                             \\ \cline{2-21} 
\multicolumn{1}{l|}{} & \multicolumn{1}{c}{Acc}   & \multicolumn{1}{c}{AP}    & \multicolumn{1}{c}{AR}    & \multicolumn{1}{c|}{AF1}   & \multicolumn{1}{c}{Acc}   & \multicolumn{1}{c}{AP}    & \multicolumn{1}{c}{AR}    & \multicolumn{1}{c|}{AF1}   & \multicolumn{1}{c}{Acc}   & \multicolumn{1}{c}{AP}    & \multicolumn{1}{c}{AR}    & \multicolumn{1}{c|}{AF1}   & \multicolumn{1}{c}{Acc}   & \multicolumn{1}{c}{AP}    & \multicolumn{1}{c}{AR}    & \multicolumn{1}{c|}{AF1}   & \multicolumn{1}{c}{Acc} & \multicolumn{1}{c}{AP} & \multicolumn{1}{c}{AR} & \multicolumn{1}{c}{AF1} \\ \hline             
Deep SVDD     &   0.333                        &     0.559                      &         0.568                  &    0.564                        &    0.714                      &      0.750                     &       0.735                    &        0.743                    &       0.385                    &     0.727                      &           0.725                &    0.726                        &          0.297                 &     0.586                      &     0.588                      &      0.587                      &    0.360                     &      0.703                  &      0.698                  &     0.700                    \\
\multicolumn{1}{c|}{AnoTrans} &    0.357                       &     0.647                      &    0.643                       &   0.645                         &       0.286                    &       0.583                    &      0.577                     &    0.580                        &     0.538                      &     0.767                      &        0.752                   &      0.759                      &       0.297                    &        0.656                   &        0.655                   &     0.655                       &        0.400                 &      0.757                  &      0.753                  &         0.755                \\
\multicolumn{1}{c|}{DCdetector} &     0.571                      &     0.541                      &     0.572                      &               0.556             &          0.571                &        0.474                   &        0.514                   &     0.493                       &       \underline{\color{blue}{0.846}}                    &       0.619                    &       0.757                    &    0.681                       &     0.220                      &     0.259                      &      0.365                     &    0.303                        &    0.640                     &   0.663                     &      0.781                  &    0.717                     \\
\multicolumn{1}{c|}{MEMTO} &     0.405                      &       0.660                    &      0.650                     &    0.655                        &       0.714                    &     0.758                      &        0.743                   &      0.750                      &     0.615                      &    0.745                       &    0.754                       &        0.750                    &    0.319                       &   0.617                        &     0.614                      &     0.616                       &   0.440                      &     0.734                   &         0.732               &       0.733                  \\
MTGFlow     &       0.500                    &        0.539                   &       0.578                   &        0.558                    &    0.714                       &     \underline{\color{blue}0.907}                      &      \underline{\color{blue}0.901}                     &         \underline{\color{blue}0.904}                   &          0.462                 &       0.682                    &      0.692                     &       0.687                     &    0.286                       &   0.603                        &   0.602     &       0.602      &      0.520                     &       0.662                     &   0.660                        &      0.661     \\
GPT4TS     &   0.476                        &     0.686                      &        0.676                   &      0.681                      &    0.429                       &     0.506                      &      0.501                     &         0.504                   &          0.462                 &       0.678                    &      0.694                     &       0.686                     &   0.330                        &      0.610                     &  0.607      &      0.608       &      0.360                     &       0.760                     &   0.759                        &      0.759     \\
TS-TCC     &     0.690                      &      0.763                     &      0.745                     &     0.754                       &      0.286                     &     0.555                      &     0.543                      &       0.549                     &     \textbf{\color{red}{1.000}}                      &      0.980                     &     0.957                      &         0.969                   &             \underline{\color{blue}{0.637}}              &      \textbf{\color{red}{0.785}}                     &     \underline{\color{blue}{0.782}}                      &        0.784                    &                 \underline{\color{blue}{0.880}}        &     \underline{\color{blue}{0.928}}                   &     \textbf{\color{red}{0.935}}                   &   \underline{\color{blue}{0.931}}                      \\
THOC     &    \underline{\color{blue}{0.762}}                       &      \underline{\color{blue}{0.822}}                     &      \underline{\color{blue}{0.808}}                     &    \underline{\color{blue}{0.815}}                        &     0.714                      &      0.782                     &      0.770                     &      0.776                      &       \textbf{\color{red}{1.000}}                    &       \underline{\color{blue}{0.984}}                    &       \underline{\color{blue}{0.958}}                    &        \underline{\color{blue}{0.971}}                    &             0.604              &     0.762                      &     0.758                      &      0.760                      &    0.880                     &          0.911              &   0.905                     &       0.908                  \\
NCAD     &   0.680                        &      0.802                     &      0.786                     &        0.794                    &     \underline{\color{blue}{0.846}}                      &     0.855                      &     0.842                     &    0.849                        &         0.714                  &     0.762                      &      0.747                     &      0.758                      &               0.593            &    0.737                       &        0.732                   &    0.735                        &  0.760                       &     0.795                   &    0.783                    &      0.789                   \\
COCA                  & 0.714                     & 0.748                     & 0.732                     & 0.740                      & 0.428                     & 0.545                     & 0.546                     & 0.545                      & \textbf{\color{red}{1.000}}                      & 0.959                     & 0.935                     & 0.946                      & \underline{\color{blue}{0.637}}                     & \underline{\color{blue}{0.769}}                     & 0.765                     & 0.767                      & 0.640                   & 0.788                  & 0.783                  & 0.785                   \\
Ours                   &  \textbf{\color{red}{0.857}}                     & \textbf{\color{red}{0.931}}                     & \textbf{\color{red}{0.910}}                     & \textbf{\color{red}{0.920}}                      & \textbf{\color{red}{1.000}}                      & \textbf{\color{red}{0.965}}                     & \textbf{\color{red}{0.948}}                     & \textbf{\color{red}{0.956}}                      & \textbf{\color{red}{1.000}}                      & \textbf{\color{red}{0.989}}                     & \textbf{\color{red}{0.959}}                     & \textbf{\color{red}{0.974}}                      & \textbf{\color{red}{0.758}}                     & 0.768                     & \textbf{\color{red}{0.808}}                     & \textbf{\color{red}{0.787}}                      & \textbf{\color{red}{0.920}}                   & \textbf{\color{red}{0.935}}                  & \underline{\color{blue}{0.932}}                  & \textbf{\color{red}{0.933}}                   \\ \hline
                      & \multicolumn{4}{c|}{Gait}                                                                                      & \multicolumn{4}{c|}{NASA}                                                                                      & \multicolumn{4}{c|}{PowerDemand}                                                                               & \multicolumn{4}{c|}{RESP}                                                                                      & \multicolumn{4}{c}{Avg}                                                                             \\ \cline{2-21} 
                      & \multicolumn{1}{c}{Acc}   & \multicolumn{1}{c}{AP}    & \multicolumn{1}{c}{AR}    & \multicolumn{1}{c|}{AF1}   & \multicolumn{1}{c}{Acc}   & \multicolumn{1}{c}{AP}    & \multicolumn{1}{c}{AR}    & \multicolumn{1}{c|}{AF1}   & \multicolumn{1}{c}{Acc}   & \multicolumn{1}{c}{AP}    & \multicolumn{1}{c}{AR}    & \multicolumn{1}{c|}{AF1}   & \multicolumn{1}{c}{Acc}   & \multicolumn{1}{c}{AP}    & \multicolumn{1}{c}{AR}    & \multicolumn{1}{c|}{AF1}   & \multicolumn{1}{c}{Acc} & \multicolumn{1}{c}{AP} & \multicolumn{1}{c}{AR} & \multicolumn{1}{c}{AF1} \\ \hline            
Deep SVDD     &    0.242                       &     0.486                      &      0.505                     &     0.495                       &    0.182                       &    0.437                       &      0.433                     &     0.435                       &      0.182                     &         0.418                  &     0.418                      &       0.418                     &        0.000                   &      0.207                     &      0.246                     &     0.225                       &     0.288                    &             0.552           &    0.558                    &      0.555                   \\
\multicolumn{1}{c|}{AnoTrans} &     0.364                      &    0.685                       &       0.682                    &            0.683                &       0.636                    &          0.821                 &          0.815                 &     0.819                       &    0.455                       &     0.663                      &     0.658                      &     0.660                       &      0.117                     &    \underline{\color{blue}{0.667}}                       &       \underline{\color{blue}{0.667}}                   &    \underline{\color{blue}{0.667}}                        &    0.348                     &   0.680                     &      0.677                  &   0.679                      \\
\multicolumn{1}{c|}{DCdetector} &     0.424                      &    0.381                       &   0.515                        &                 0.438           &         \underline{\color{blue}{0.909}}                  &      0.608                     &     \underline{\color{blue}{0.931}}                      &    0.736                        &         0.455                  &       0.491                    &      0.493                     &    0.492                        &       0.117                    &      0.517                     &     0.682                      &          0.588                  &     0.424                    &    0.431                    &      0.538                  &        0.479                 \\
\multicolumn{1}{c|}{MEMTO} &    0.364                       &    0.674                       &      0.671                     &              0.673              &      0.364                     &     0.662                      &      0.665                     &    0.664                        &      0.455                     &     0.683                      &     0.677                      &     0.680                       &     0.059                      &       0.598                    &      0.598                     &        0.598                    &     0.368                    &    0.658                    &       0.654                 &        0.656                 \\
MTGFlow     &    0.364                       &       0.574                    &     0.571                      &      0.572                      &    0.364                       &     0.747                      &      0.751                     &         0.749                   &          0.182                 &       0.573                    &      0.575                     &       0.574                     &   0.235                        &      0.512                     &  0.514      &      0.513       &        0.372                 &       0.606                     &    0.612                      &    0.608       \\
GPT4TS     &   0.212                        &    0.463                       &      0.460                     &      0.461                      &       0.364                    &     0.850                      &        0.848                   &       0.849                     &             0.182              &    0.597                       &    0.598                       &       0.598                     &              0.353             &        0.544                   &   0.543       &  0.544       &   0.348      &     0.624      &    0.622       &  0.623   \\ 
TS-TCC     &    0.697                       &    0.798                       &      0.790                     &      0.794                      &       0.364                    &      0.512                     &       0.508                    &         0.511                  &       \underline{\color{blue}{0.545}}                    &    0.767                       &    0.759                       &       0.763                     &            \underline{\color{blue}{0.412}}               &          0.561                 &      0.560                    &     0.560                       &     0.656                    &       0.773                 &     0.766                   &    0.770                     \\
THOC     &    0.636                       &     0.788                      &  0.780                         &      0.784                      &    \underline{\color{blue}{0.909}}                       &      \underline{\color{blue}{0.902}}                     &      0.891                     &     \underline{\color{blue}{0.896}}                       &       0.455                    &     \underline{\color{blue}{0.777}}                      &    \underline{\color{blue}{0.772}}                       &       \underline{\color{blue}{0.775}}                     &            0.294              &     0.382                      &     0.395                      &     0.389                       &    \underline{\color{blue}{0.671}}                     &    \underline{\color{blue}{0.783}}                    &    \underline{\color{blue}{0.777}}                    &     \underline{\color{blue}{0.780}}                    \\
NCAD     &   \underline{\color{blue}{0.848}}                        &    \underline{\color{blue}{0.864}}                       &     \underline{\color{blue}{0.852}}                      &      \underline{\color{blue}{0.858}}                      &     0.818                      &      0.869                     &     0.853                      &      0.861                      &       \underline{\color{blue}{0.545}}                    &     0.724                      &       0.723                    &     0.723                       &             0.353            &     0.613                      &       0.612                    &       0.613                     &    0.663                    &    0.772                    &    0.763                    &  0.767                       \\
COCA                  & 0.545                     & 0.703                     & 0.694                     & 0.699                      & 0.818                     & 0.849                     & 0.833                     & 0.841                      & 0.364                     & 0.633                     & 0.631                     & 0.632                      & 0.235                     & 0.563                     & 0.562                     & 0.562                      &  0.620                    &  0.746                      &   0.738                     &   0.742                      \\
Ours                   & \multicolumn{1}{c}{\textbf{\color{red}{0.878}}} & \multicolumn{1}{c}{\textbf{\color{red}{0.891}}} & \multicolumn{1}{c}{\textbf{\color{red}{0.852}}} & \multicolumn{1}{c|}{\textbf{\color{red}{0.871}}} & \multicolumn{1}{c}{\textbf{\color{red}{1.000}}} & \multicolumn{1}{c}{\textbf{\color{red}{0.969}}} & \multicolumn{1}{c}{\textbf{\color{red}{0.953}}} & \multicolumn{1}{c|}{\textbf{\color{red}{0.961}}} & \multicolumn{1}{c}{\textbf{\color{red}{0.818}}} & \multicolumn{1}{c}{\textbf{\color{red}{0.888}}} & \multicolumn{1}{c}{\textbf{\color{red}{0.884}}} & \multicolumn{1}{c|}{\textbf{\color{red}{0.886}}} & \multicolumn{1}{c}{\textbf{\color{red}{0.471}}} & \multicolumn{1}{c}{\textbf{\color{red}{0.736}}} & \multicolumn{1}{c}{\textbf{\color{red}{0.736}}} & \multicolumn{1}{c|}{\textbf{\color{red}{0.736}}} & \multicolumn{1}{c}{\textbf{\color{red}{0.820}}}    & \multicolumn{1}{c}{\textbf{\color{red}{0.857}}}   & \multicolumn{1}{c}{\textbf{\color{red}{0.860}}}   & \multicolumn{1}{c}{\textbf{\color{red}{0.858}}}    \\ \hline
\end{tabular}}
\caption{Overall results on the UCR datasets.}\label{tab:ucr}
\vspace{-1.5em}
\end{table*}

\subsection{Model Training}
We aim to distinguish anomaly representations of the student and teacher networks. Given the absence of anomaly labels under unsupervised settings, we propose a data augmentation-based training strategy.

Firstly, we apply data augmentation to generate synthetic anomalies. To ensure that the general pattern of the sample does not change significantly, we randomly select a segment from the entire time window and apply augmentation to this segment. We use augmentation methods including jittering \cite{iwana2021empirical}, scaling \cite{wang2023deep}, and warping \cite{sun2023test}. The synthetic sample based on the original sample $\mathbf{w}_i$ is denoted as $\mathbf{w}_i^a$.

Next, both original and synthetic samples are fed into the teacher and student networks. The generated representation pairs of original and synthetic samples are denoted as  $(z_i, c_i)$ and $(z_i^a, c_i^a)$, respectively.
We push away the representation pair of the synthetic sample while pulling together that of the original samples. This knowledge distillation loss is calculated based on Eq.~\ref{eq:1}:
\begin{equation}
\mathcal{L}_{kd} = \frac{1}{N}\sum\limits_{i=1}^{N} {\Vert z_{i}-c_{i} \Vert}_{2}^{2} - \log(1-\exp(-{\Vert z_{i}^{a}-c_{i}^{a} \Vert}_{2}^{2})),
\end{equation}
where $N$ is the total number of training samples. 

Additionally, to enable the teacher network to focus on more general patterns and produce representations robust to noise, we consider the teacher's representations of the original and synthetic samples as positive pairs and aim to minimize the distance between them. This negative-sample-free contrastive loss \cite{wang2023deep} is defined as:
\begin{equation}
\mathcal{L}_{ce}=\frac{1}{N}\sum\limits_{i=1}^{N}-\frac{c_i}{\Vert c_i \Vert_{2}} \cdot \frac{c_i^{a}}{\Vert c_i^{a} \Vert_{2}}.\label{eq:5}
\end{equation} Integrating this contrastive loss, 
the complete loss function is defined as:
\begin{equation}
\mathcal{L}_{total}=\mathcal{L}_{kd} +  \lambda  \mathcal{L}_{ce},
\end{equation}
where $\lambda$ is a hyperparameter that controls the weight of knowledge distillation loss and contrastive loss.

During the testing phase, we calculate the anomaly score of the given time window $\mathbf{w}$ according to:
\begin{equation}
A(\mathbf{w}) = {\Vert \phi(\mathbf{w}) - \varphi(\mathbf{w})  \Vert}_2^2.
\end{equation}

\section{Experiment}




\subsection{Datasets}
\textbf{UCR Anomaly Archive (UCR).} Wu \emph{et~al.} \cite{wu2021current} identify several flaws in previously used benchmarks and introduce the UCR time-series anomaly archive. This archive comprises 250 diverse univariate time series signals spanning various domains. Following \cite{goswami2022unsupervised}, we partition the complete UCR archive into 9 separate datasets based on the domain to which each signal belongs, (1) Arterial Blood Pressure, ABP, (2) Acceleration, (3) Air Temperature, (4) Electrocardiogram, ECG, (5) Electrical Penetration Graph, EPG, (6) Gait, (7) NASA, (8) Power Demand, and (9) Respiration, RESP.

\noindent
\textbf{Previous benchmarks.} We also evaluate our method on 6 previously commonly used multivariate datasets, including (1) SMD: It contains five-week-long metrics from 28 server machines in an internet company. (2) MSL: It records the conditions of sensors from the Mars rover provided by NASA (3) SMAP: It is also collected by NASA and gathers the soil samples and telemetry information used by the Mars rover. (4) PSM: Provided by eBay, it records 25-dimensional metrics from the server machines. (5) NIPS-TS-GECCO: It is a drinking water quality dataset collected through the Internet of Things.  (6) NIPS-TS-SWAN: It contains space weather data from Harvard Dataverse. 

In total, we use 15 datasets (9 univariate datasets and 6 multivariate datasets) from various domains for evaluation.

\subsection{Settings}
\paragraph{Evaluation Metrics.} Traditional metrics for anomaly detection include precision, recall, and F1 score. Point adjustment and revised point adjustment are also utilized to post-process the metrics. However, previous works have demonstrated that these metrics might lead to an overestimation of method performance \cite{kim2022towards}. In this paper, we adopt affiliation metrics to assess the performance from an event-wise perspective \cite{huet2022local}. Precision, recall, and F1-score are calculated based on the affiliation between ground truth and prediction sets. For UCR datasets, we also employ accuracy as a metric \cite{wang2023deep}, indicating the probability of correctly predicting subdatasets.

\paragraph{Hyperparameters.}
The baselines are implemented based on the hyperparameters reported in previous literature. For our model, we use the pretrained GPT2 with 6 layers as our teacher network. Regarding the student network, we use a pool with 32 prototypes and an attention mechanism with an intermediate dimension of 64 and a head number of 8. During the training stage, we utilize an Adam optimizer with a learning rate of 0.0001 and a batch size of 32. All experiments are conducted on a single RTX 3090. 

\paragraph{Baselines.}
We compare our method with 10 baselines for evaluation, including the one-class classification-based methods: DeepSVDD \cite{ruff2018deep}, THOC \cite{shen2020timeseries}, NCAD \cite{carmona2021neural}; the density estimation-based method: MTGFlow \cite{zhou2023detecting}; the self-supervised methods: AnoTrans \cite{xu2021anomaly}, MEMTO \cite{song2023memto}, TS-TCC \cite{eldele2021time}, COCA \cite{wang2023deep}, DCdetector \cite{yang2023dcdetector}; the LLM-based method: GPT4TS \cite{zhou2023one}.

\begin{figure*}[t]
\centering
\includegraphics[width=1\textwidth]{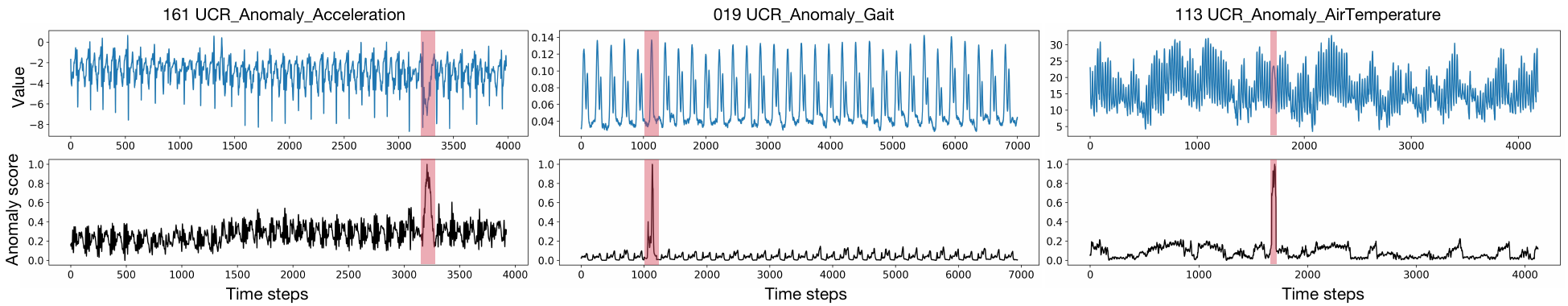}
\caption{Case studies of anomaly score visualization. }\label{fig:v1}
\end{figure*}

\begin{table*}[t]
\scalebox{0.9}{
\begin{tabular}{c|lllllllllllll}
\hline
\multicolumn{1}{l|}{} & \multicolumn{3}{c}{SMD}                                                            & \multicolumn{3}{c}{MSL}                                                            & \multicolumn{3}{c}{SMAP}                                                           & \multicolumn{3}{c}{PSM}                                                            & Avg                       \\ \cline{2-14} 
\multicolumn{1}{l|}{} & \multicolumn{1}{c}{P}     & \multicolumn{1}{c}{R}     & \multicolumn{1}{c|}{F1}    & \multicolumn{1}{c}{P}     & \multicolumn{1}{c}{R}     & \multicolumn{1}{c|}{F1}    & \multicolumn{1}{c}{P}     & \multicolumn{1}{c}{R}     & \multicolumn{1}{c|}{F1}    & \multicolumn{1}{c}{P}     & \multicolumn{1}{c}{R}     & \multicolumn{1}{c|}{F1}    & \multicolumn{1}{c}{F1}    \\ \hline
LOF                   & \multicolumn{1}{c}{0.563} & \multicolumn{1}{c}{0.399} & \multicolumn{1}{c|}{0.467} & \multicolumn{1}{c}{0.477} & \multicolumn{1}{c}{0.853} & \multicolumn{1}{c|}{0.612} & \multicolumn{1}{c}{0.589} & \multicolumn{1}{c}{0.563} & \multicolumn{1}{c|}{0.576} & \multicolumn{1}{c}{0.579} & \multicolumn{1}{c}{0.905} & \multicolumn{1}{c|}{0.706} & \multicolumn{1}{c}{0.609} \\
OC-SVM                & 0.443                     & 0.767                     & \multicolumn{1}{l|}{0.562} & 0.598                     & 0.869                     & \multicolumn{1}{l|}{0.708} & 0.539                     & 0.591                     & \multicolumn{1}{l|}{0.563} & 0.628                     & 0.809                     & \multicolumn{1}{l|}{0.707} & 0.603                     \\
Isolation Foreset     & 0.423                     & 0.733                     & \multicolumn{1}{l|}{0.536} & 0.539                     & 0.865                     & \multicolumn{1}{l|}{0.665} & 0.524                     & 0.591                     & \multicolumn{1}{l|}{0.555} & 0.761                     & 0.925                     & \multicolumn{1}{l|}{0.835} & 0.612                     \\
MMPCACD               & 0.712                     & 0.793                     & \multicolumn{1}{l|}{0.750} & 0.814                     & 0.613                     & \multicolumn{1}{l|}{0.700} & 0.886                     & 0.758                     & \multicolumn{1}{l|}{0.817} & 0.763                     & 0.784                     & \multicolumn{1}{l|}{0.773} & 0.757                     \\
DAGMM                 & 0.673                     & 0.499                     & \multicolumn{1}{l|}{0.573} & 0.896                     & 0.639                     & \multicolumn{1}{l|}{0.746} & 0.865                     & 0.567                     & \multicolumn{1}{l|}{0.685} & 0.935                     & 0.700                     & \multicolumn{1}{l|}{0.801} & 0.702                     \\
Deep-SVDD             & 0.785                     & 0.797                     & \multicolumn{1}{l|}{0.791} & 0.919                     & 0.766                     & \multicolumn{1}{l|}{0.836} & 0.899                     & 0.560                     & \multicolumn{1}{l|}{0.690} & 0.954                     & 0.865                     & \multicolumn{1}{l|}{0.907} & 0.810                     \\
THOC                  & 0.798                     & 0.910                     & \multicolumn{1}{l|}{0.850} & 0.885                     & 0.910                     & \multicolumn{1}{l|}{0.897} & 0.921                     & 0.893                     & \multicolumn{1}{l|}{0.907} & 0.881                     & 0.910                     & \multicolumn{1}{l|}{0.895} & 0.880                     \\
LSTM-VAE              & 0.758                     & 0.901                     & \multicolumn{1}{l|}{0.823} & 0.855                     & 0.799                     & \multicolumn{1}{l|}{0.826} & 0.922                     & 0.678                     & \multicolumn{1}{l|}{0.781} & 0.736                     & 0.899                     & \multicolumn{1}{l|}{0.810} & 0.812                     \\
BeatGAN               & 0.729                     & 0.841                     & \multicolumn{1}{l|}{0.781} & 0.898                     & 0.854                     & \multicolumn{1}{l|}{0.875} & 0.924                     & 0.559                     & \multicolumn{1}{l|}{0.696} & 0.903                     & 0.938                     & \multicolumn{1}{l|}{0.920} & 0.802                     \\
OmniAnomaly           & 0.837                     & 0.868                     & \multicolumn{1}{l|}{0.852} & 0.890                     & 0.864                     & \multicolumn{1}{l|}{0.877} & 0.925                     & 0.820                     & \multicolumn{1}{l|}{0.869} & 0.814                     & 0.843                     & \multicolumn{1}{l|}{0.828} & 0.847                     \\
InterFusion           & 0.870                     & 0.854                     & \multicolumn{1}{l|}{0.862} & 0.813                     & 0.927                     & \multicolumn{1}{l|}{0.866} & 0.898                     & 0.885                     & \multicolumn{1}{l|}{0.891} & 0.836                     & 0.835                     & \multicolumn{1}{l|}{0.835} & 0.857                     \\
Anomaly Transformer   & 0.880                     & 0.947                     & \multicolumn{1}{l|}{0.912} & 0.911                     & 0.901                     & \multicolumn{1}{l|}{0.906} & \underline{\color{blue}{0.940}}                     & 0.985                     & \multicolumn{1}{l|}{0.962} & 0.968                     & 0.986                     & \multicolumn{1}{l|}{0.977} & 0.936                     \\
DCdetector            & 0.836                     & 0.911                     & \multicolumn{1}{l|}{0.872} & \underline{\color{blue}{0.937}}                     & \textbf{\color{red}{0.997}}                     & \multicolumn{1}{l|}{\textbf{\color{red}{0.966}}} & \textbf{\color{red}{0.956}}                     & 0.989                     & \multicolumn{1}{l|}{\textbf{\color{red}{0.970}}} & 0.971                     & 0.987                     & \multicolumn{1}{l|}{0.979} & 0.946                     \\ 
MEMTO                 & \underline{\color{blue}{0.891}}                     & \textbf{\color{red}{0.984}}                     & \multicolumn{1}{l|}{\underline{\color{blue}{0.935}}} & 0.921                     & 0.968                     & \multicolumn{1}{l|}{0.944} & 0.938                     & \underline{\color{blue}{0.996}}                     & \multicolumn{1}{l|}{\underline{\color{blue}{0.966}}} & \underline{\color{blue}{0.975}}                     & \underline{\color{blue}{0.992}}                     & \multicolumn{1}{l|}{\underline{\color{blue}{0.983}}} & \underline{\color{blue}{0.957}}                     \\ \hline
Ours                   & \textbf{\color{red}{0.937}}                     & \underline{\color{blue}{0.979}}                     & \multicolumn{1}{l|}{\textbf{\color{red}{0.958}}} & \textbf{\color{red}{0.944}}                     & \underline{\color{blue}{0.969}}                     & \multicolumn{1}{l|}{\underline{\color{blue}{0.956}}} & 0.934                     & \textbf{\color{red}{0.998}}                     & \multicolumn{1}{l|}{0.965} & \textbf{\color{red}{0.996}}                     & \textbf{\color{red}{0.998}}                     & \multicolumn{1}{l|}{\textbf{\color{red}{0.997}}} & \textbf{\color{red}{0.969}}                     \\ \hline
\end{tabular}}
\caption{Overall results on SMD, MSL, SMAP, and PSM.}\label{tab:smd}
\vspace{-1em}
\end{table*}

\begin{table}[t]
\scalebox{0.9}{
\begin{tabular}{c|llllll}
\hline
\multicolumn{1}{l|}{} & \multicolumn{3}{c}{NIPS-TS-GECCO}                                                  & \multicolumn{3}{c}{NIPS-TS-SWAN}                                                  \\ \cline{2-7} 
\multicolumn{1}{l|}{} & \multicolumn{1}{c}{P}     & \multicolumn{1}{c}{R}     & \multicolumn{1}{c|}{F1}    & \multicolumn{1}{c}{P}     & \multicolumn{1}{c}{R}     & \multicolumn{1}{c}{F1}    \\ \hline
OCSVM                 & \multicolumn{1}{c}{0.021} & \multicolumn{1}{c}{0.341} & \multicolumn{1}{c|}{0.040} & \multicolumn{1}{c}{0.193} & \multicolumn{1}{c}{0.001} & \multicolumn{1}{c}{0.001} \\
MatrixProfile         & 0.046                     & 0.185                     & \multicolumn{1}{l|}{0.074} & 0.167                     & 0.175                     & 0.171                     \\
GBRT                  & 0.175                     & 0.140                     & \multicolumn{1}{l|}{0.156} & 0.447                     & 0.375                     & 0.408                     \\
LSTM-RNN              & 0.343                     & 0.275                     & \multicolumn{1}{l|}{0.305} & 0.527                     & 0.221                     & 0.312                     \\
Autoregression        & 0.392                     & 0.314                     & \multicolumn{1}{l|}{0.349} & 0.421                     & 0.354                     & 0.385                     \\
IForest               & \underline{\color{blue}{0.439}}                     & 0.353                     & \multicolumn{1}{l|}{0.391} & 0.569                     & \underline{\color{blue}{0.598}}                     & 0.583                     \\
AutoEncoder           & 0.424                     & 0.340                     & \multicolumn{1}{l|}{0.377} & 0.497                     & 0.522                     & 0.509                     \\
AnomalyTrans          & 0.257                     & 0.285                     & \multicolumn{1}{l|}{0.270} & 0.907                     & 0.474                     & 0.623                     \\
MTGFlow          & 0.333                     & 0.125                     & \multicolumn{1}{l|}{0.182} & \textbf{\color{red}{1.000}}                     & 0.494                     & 0.662                     \\
DCdetector            & 0.383                     & \underline{\color{blue}{0.597}}                     & \multicolumn{1}{l|}{\underline{\color{blue}{0.466}}} & \underline{\color{blue}{0.955}}                     & 0.596                     & \underline{\color{blue}{0.734}}                     \\ \hline
Ours                   & \textbf{\color{red}{0.511}}                     & \textbf{\color{red}{0.793}}                     & \multicolumn{1}{l|}{\textbf{\color{red}{0.620}}} & 0.873                     & \textbf{\color{red}{0.745}}                     & \textbf{\color{red}{0.804}}                     \\ \hline
\end{tabular}}
\caption{Overall results on the NIPS benchmark.}\label{tab:nips}
\vspace{-1.5em}
\end{table}

\subsection{Model Comparison}

Table~\ref{tab:ucr} presents the performance of all methods on the UCR datasets. Our method consistently achieves the highest accuracy and affiliated F1 (AF1) score across all domains, demonstrating an average improvement of 22.2\% in accuracy and 10.0\% in AF1 compared to the second-best method. Notably, our method detects 82\% of anomalies in total. Furthermore, three key observations can be made. First, while conventional DeepSVDD struggles to deliver satisfactory performance, one-class classification methods like THOC and NCAD show great potential in time series anomaly detection when they incorporate temporal information within the time window. Second, contrastive learning-based methods (TS-TCC and COCA) outperform reconstruction-based methods (AnoTrans, MEMTO), suggesting that approaching anomaly detection from a representation perspective has advantages over using original signals. Third, the direct application of LLM (GPT4TS) falls short of expectations. This is attributed to the overstrong ability of the LLM-based autoencoder to reconstruct even abnormal time series, resulting in poor performance. Our method overcomes this limitation by introducing a small student network to extract features without overgeneralization. 

We extend the evaluation of our method to multivariate datasets and present results in Table~\ref{tab:smd} and Table~\ref{tab:nips}. Notably, most previous methods report results after employing the point adjustment strategy. To ensure a fair comparison, we also include the adjusted metrics for these datasets. For datasets such as SMD, MSL, SMAP, and PSM which contain a large number of obvious anomalies \cite{wu2021current}, our method demonstrates comparable performance with SOTA methods. In the case of NIPS-TS-GECCO and NIPS-TS-SWAN, which present diverse and challenging anomalies, our method outperforms DCdetector significantly, achieving a 62\% F1 score compared to 47\% on GECCO and an 80\% F1 score compared to 73\% on SWAN.




\begin{figure}[t]
\centering
\includegraphics[width=0.5\textwidth]{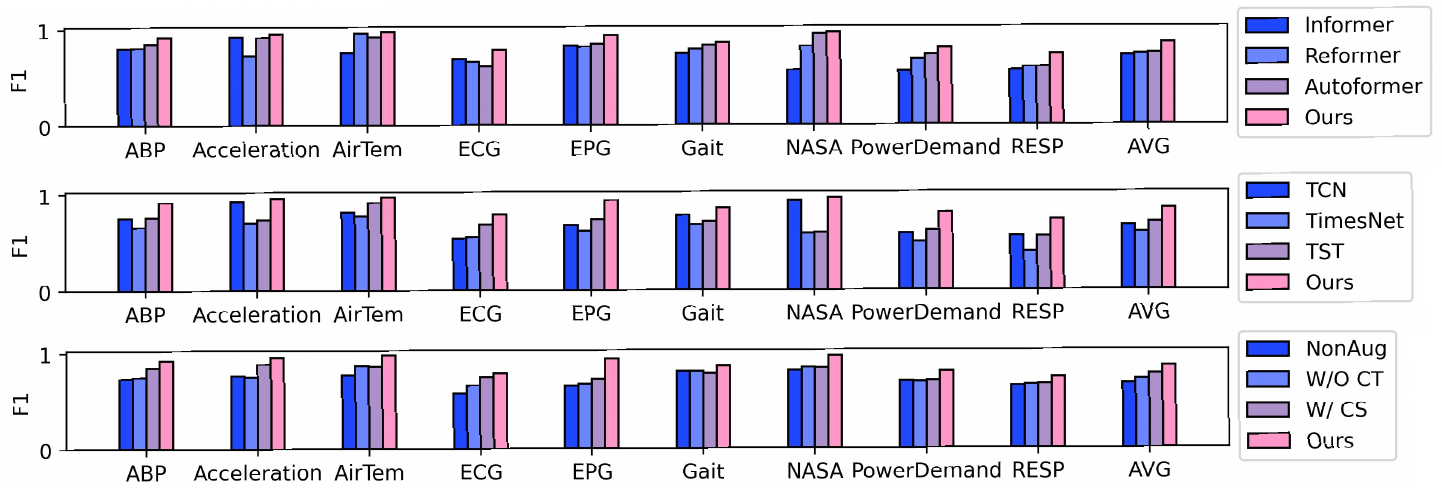}
\caption{Ablation studies. \textbf{Top:} It depicts the performance of the model with different center predictors. \textbf{Middle:} It depicts the performance of the models with different projectors. \textbf{Bottom:} It depicts the performance of different training strategies. }\label{fig:abla}
\vspace{-1.5em}
\end{figure}

\begin{figure*}[ht]
\centering
\subfigure[Window sizes.]{\includegraphics[width=0.19\textwidth]{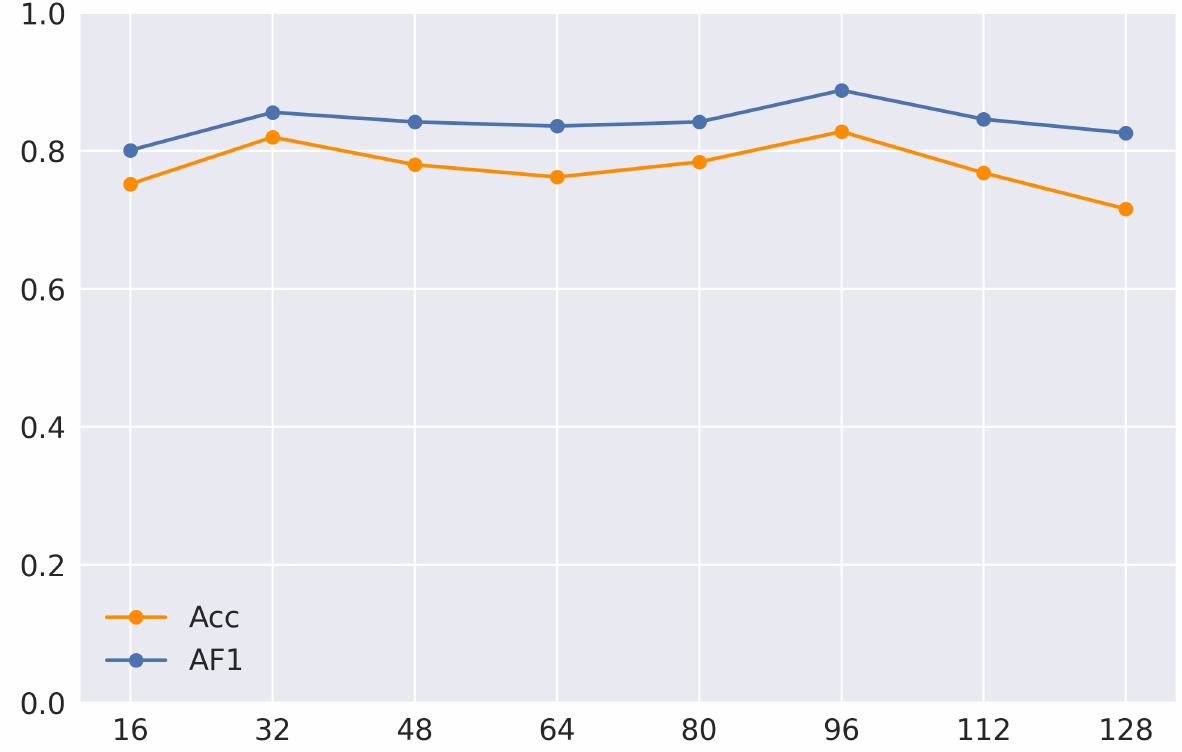}\label{fig:para1}}
\hfill
\subfigure[Augmentations.]{\includegraphics[width=0.19\textwidth]{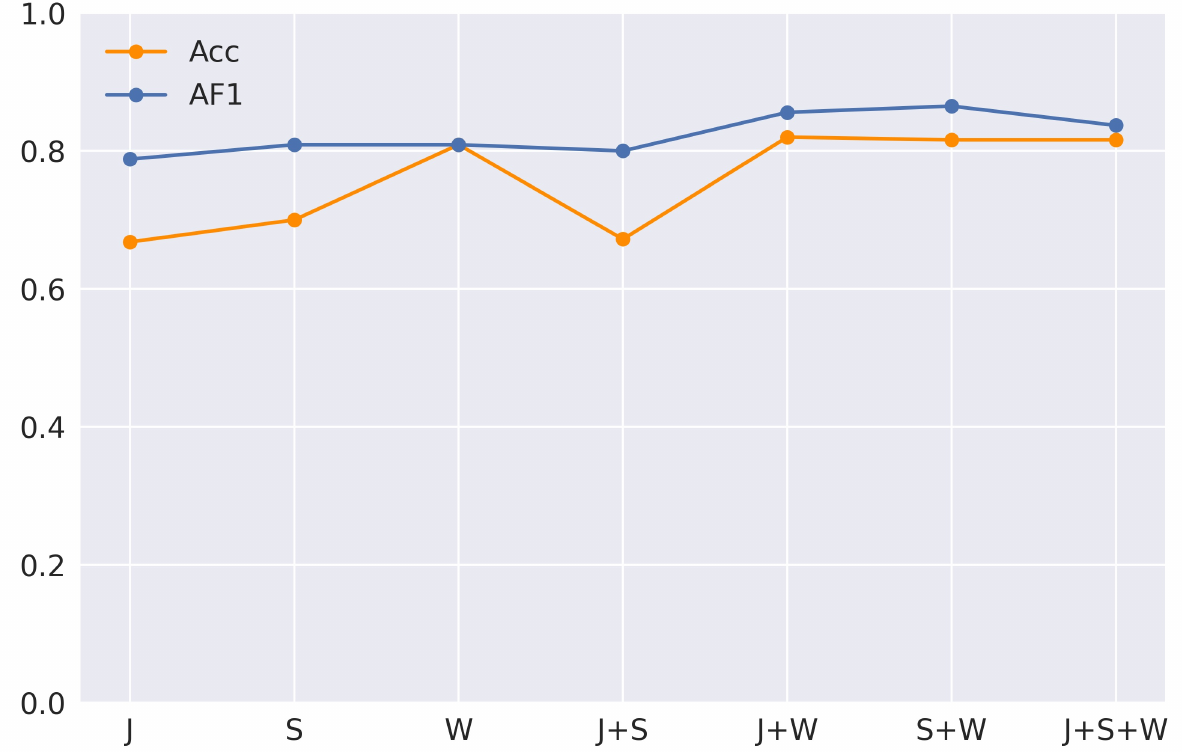}\label{fig:para2}}
\hfill
\subfigure[GPT2 layers.]{\includegraphics[width=0.19\textwidth]{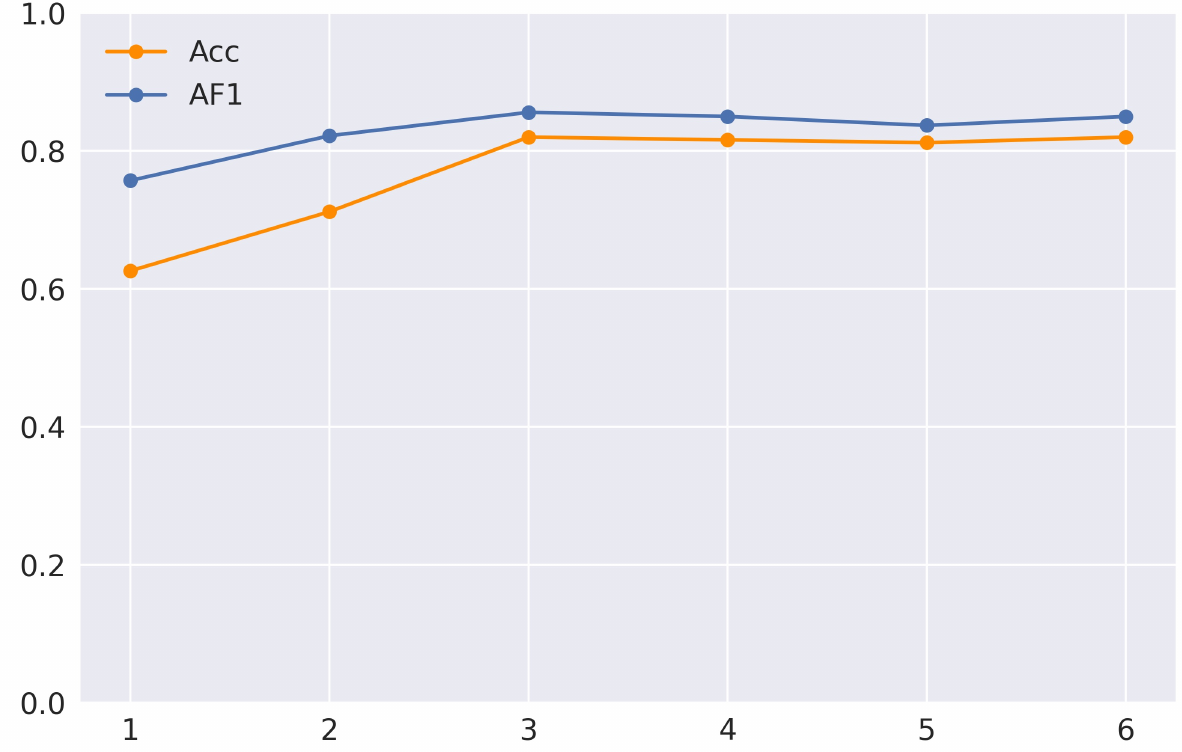}\label{fig:para3}}
\hfill
\subfigure[Student layers.]{\includegraphics[width=0.19\textwidth]{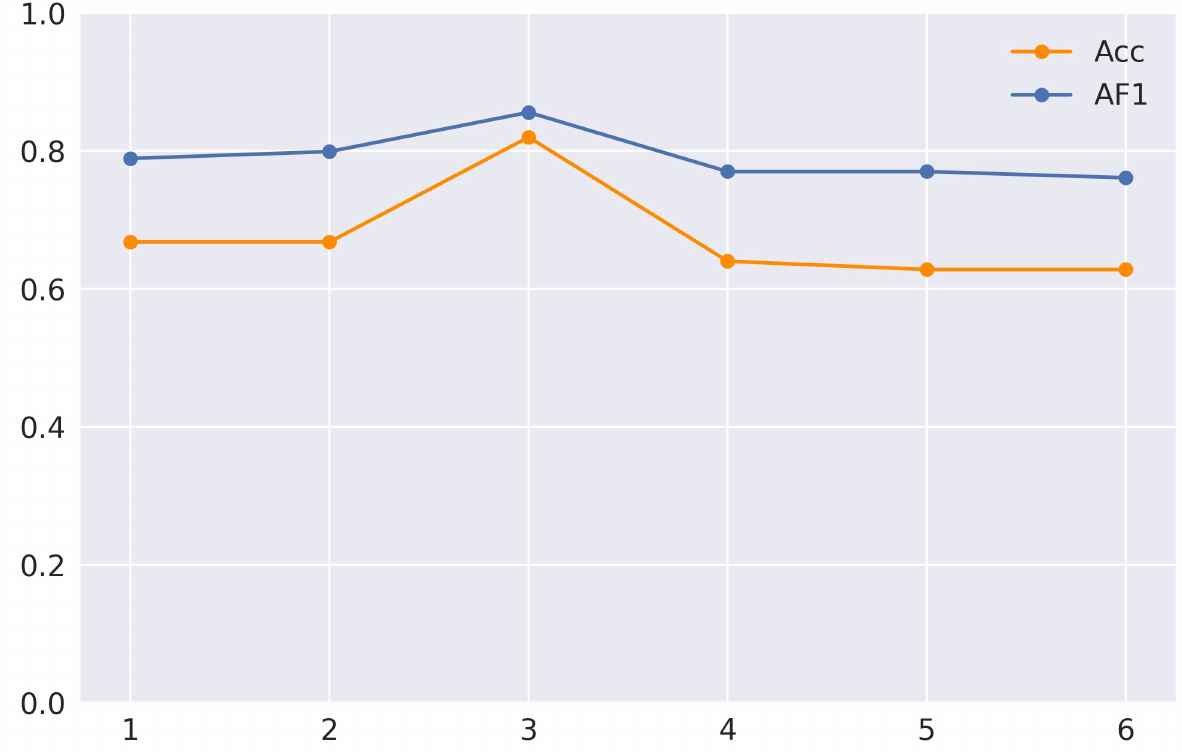}\label{fig:para4}}
\hfill
\subfigure[Prototype sizes.]{\includegraphics[width=0.19\textwidth]{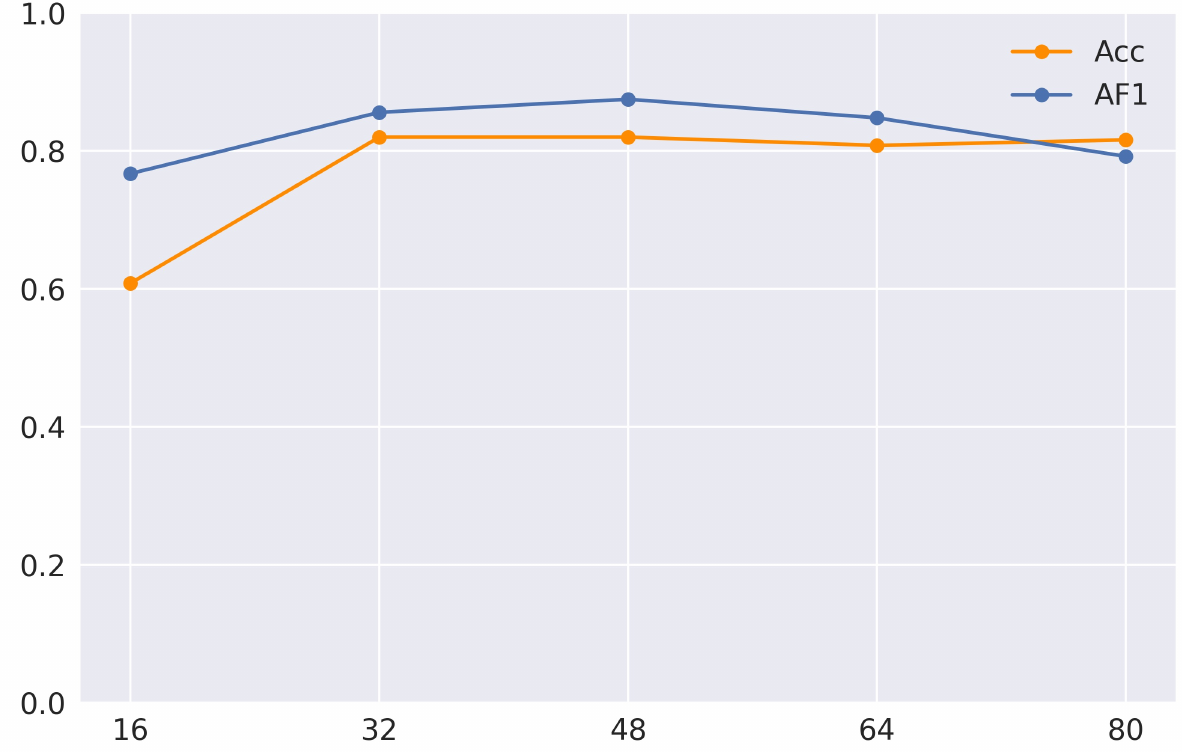}\label{fig:para5}}
\caption{Parameter sensitivity studies of main hyperparameters in AnomalyLLM.}\label{fig:para}
\end{figure*}

\subsection{Model Analysis}
\paragraph{Anomaly score visualization.}
We present three case studies in Fig.~\ref{fig:v1} to illustrate the functionality of our method. Anomaly scores are normalized and plotted for each sub-dataset. Anomalies are identified by assigning a high anomaly score, indicative of the disparity between the output of the student network and the LLM-based teacher network. In the UCR 113 dataset, the original signals exhibit non-stationary characteristics, with varying mean values across different stages. Our method demonstrates robustness to this domain shift and successfully identifies the most anomalous segment, characterized by a distinct shape.


\paragraph{Ablation study.}
In this section, our goal is to investigate the role of each component in our method.

1) LLM-based teacher network: To assess the significance of the LLM in representation generation, we replace it with three main transformer-based blocks (Informer, Reformer, and Autoformer). The blocks are trained from scratch on each dataset. The results reveal that the F1-score of variants with these replaced blocks fluctuates by less than 2.6\%, with an average AF1 score of 73.6\%. Our method utilizes GPT2 with pretrained parameters and only fine-tunes the positional embedding and layer normalization, resulting in an enhanced AF1 score of 12\%. The reason for this improvement can be ascribed to the fact that LLM has been pretrained by large NLP datasets, enabling it to capture temporal sequence patterns that can be shared by time series after fine-tuning, as demonstrated in previous research \cite{zhou2023one}. 

2) Prototype-based student network: To validate the effectiveness of our student network, we experiment with different choices (TCN, TimesNet, and TST). TCN is a classical 1D CNN-based feature extractor commonly used in time series analysis and has been applied in previous one-class classification methods such as DeepSVDD, THOC, and NCAD. TimesNet, proposed by \cite{wu2022timesnet}, rearranges time series data as a 2D tensor using Fast Fourier Transform (FFT) and leverages 2D CNN to extract features. TST is a SOTA Transformer-based feature extractor proposed by \cite{nie2022time}. The results show that TST alone does not significantly improve upon TCN, as it applies the attention mechanism like the teacher network does, which can easily result in similar representations of these two networks. In contrast, we incorporate prototypes into TST, enabling it to focus more on those historical frequent patterns and resulting in a notable improvement of 14.7\% in AF1 score.  

3) Training strategy: Finally, we explore different training strategies. The first strategy 'NonAug' excludes augmentation techniques and trains the model solely using original data, resulting in the poorest performance. The second strategy 'W/O CT' incorporates data augmentation but omits the contrastive loss between the teacher networks' representations in Eq.~\ref{eq:5}. This strategy improves AF1 by 4.2\% but still falls behind our strategy. We force the teacher network to learn the representations that are robust to the noises by employing a contrastive regularization term.  The third strategy 'W/ CS' introduces an additional contrastive loss aiming to maximize the discrepancy between the student network's representations of the original and augmented samples. This additional loss leads to an 8.0\% reduction in AF1 score, suggesting that our loss, which prioritizes the discrepancy between the representations of the student and teacher network, is more effective.

\paragraph{Parameter Sensitivity.}
In this part, we discuss the relationship between model performance and hyperparameters. Fig.~\ref{fig:para1} depicts the performance under various window sizes, showing that the performance remains relatively stable as the window size increases, maintaining an accuracy of over 72\% and an AF1 score of over 80\%. 
Fig.~\ref{fig:para2} showcases the performance with different augmentation methods (Jittering, scaling, and warping). We can find that the introduction of warping leads to an increase in performance. The performance under different numbers of layers in GPT2 is shown in Fig.~\ref{fig:para3}. The performance stabilizes when the number is above 3, indicating that GPT2 with 3 frozen layers is sufficient to describe the distribution of time series signals. Fig.~\ref{fig:para4} shows the results under different numbers of layers in the projector. It can be seen that the performance reaches its peak at 3. Finally, we find that the 48 prototype provides the most useful information for our projector to extract the representations, as shown in Fig.~\ref{fig:para5}. Complete results are reported in the appendix.



\section{Conclusion}
In this paper, we propose the first knowledge distillation-based TSAD method, named AnomalyLLM. Anomaly scores are determined by the representation discrepancy between the student and teacher networks. The teacher network is fine-tuned from a pretrained LLM to generate generalizable representations for time series signals when only limited samples are available. The student network incorporates prototypical signals to produce more domain-specific representations. Besides, we propose a data augmentation-based training strategy to enhance the representation gap on anomalous samples. AnomalyLLM surpasses SOTA approaches on 9 univariate datasets and 6 multivariate datasets, highlighting the remarkable potential of knowledge distillation and LLM in time series anomaly detection. Future research should explore lightweight versions of the teacher network within our framework, tailored for deployment in scenarios with limited computational and memory resources.
 




\appendix

\section{Experiment Setups}
\paragraph{Datasets.} We evaluate our method on nine univariate datasets from the UCR archive \cite{wu2021current}, including: (1) ABP, (2) Acceleration, (3) AirTemperature, (4) ECG, (5) EPG, (6) Gait, (7) NASA, (8) PowerDemand, (9) RESP, and six multivariate datasets including (1) SMD, (2) MSL, (3) SMAP, (4) PSM, (5) GECCO, (6) SWAN. The details are introduced in Table~\ref{tab:de1} and Table~\ref{tab:de2}.

\begin{table}[h]
\centering
\scalebox{0.87}{
\begin{tabular}{c|cccc}
\hline
Dataset        &  Train    &  Test       & Anomaly Ratio  \\  \hline
ABP         & 1036746         & 1841461       &  0.53\%   \\
Acceleration       & 38400         & 62337       &  2.45\%   \\
AirTemperature         & 52000         & 54392       &  2.86\%   \\
ECG         & 1795083         & 6047314       & 0.53\%   \\
EPG       & 119000      & 410415     &  1.29\%    \\
Gait      & 1157571     & 2784520     &   0.43\%   \\
NASA        & 38500     & 86296     &   2.35\%   \\
PowerDemand       & 197149     & 311629    &   0.81\%   \\
RESP       & 868000   & 2452953     & 0.19\%  \\ \hline
\end{tabular}}
\caption{Details of univariate datasets}\label{tab:de1}
\end{table}

\begin{table}[h]
\centering
\scalebox{0.9}{
\begin{tabular}{c|cccc}
\hline
Dataset       & Dimension    &  Train & Test   & Anomaly Ratio \\  \hline
SMD     & 38      & 708377         & 708393       &  4.16\%   \\
MSL   & 55      & 58317         & 73729       &  10.50\%   \\
SMAP     & 25      & 135183         & 427617       &  12.79\%   \\
PSM     & 25      & 132481         & 87841        & 27.75\%  \\
GECCO     & 10   & 60000      & 60000     &  1.25\%    \\
SWAN     & 39    & 69260     & 69261     &   23.80\%   \\ \hline
\end{tabular}}
\caption{Details of multivariate datasets}\label{tab:de2}
\end{table}

\paragraph{Metrics.} We evaluate the model from the event-wise perspective \cite{huet2022local}. Original metrics such as precision, recall, and F1-score are affiliated to reward those predictions close to the long abnormal events. Affiliated precision and recall are calculated according to:
\begin{equation}
\begin{aligned}
AP & =\frac{1}{\left|\hat{\mathrm{Y}}^{\prime}\right|} \int_{\hat{y}^{\prime} \in \hat{\mathrm{Y}}^{\prime}} \bar{F}_{\text {precision }}\left(\operatorname{dist}\left(\hat{y}^{\prime}, \mathcal{A}\right)\right) d y^{\prime} \\
AR & =\frac{1}{|\mathcal{A}|} \int_{a \in \mathcal{A}} \bar{F}_{\text {recall }}\left(\operatorname{dist}\left(a, \hat{Y}^{\prime}\right)\right) d a
\end{aligned}
\end{equation}
where $\hat{Y}^{\prime}$ represents the predicted anomalies, and  $\operatorname{dist}\left(\hat{y}^{\prime}, \mathcal{A}\right)=\min _{a \in \mathcal{A}}\left|\hat{y}^{\prime}-a\right|, \operatorname{dist}\left(a, \hat{Y}^{\prime}\right)=\min _{\hat{y}^{\prime} \in \hat{Y}^{\prime}}\left|a-\hat{y}^{\prime}\right|$ represent the distance between predicted and ground truth anomalies.  $\bar{F}_{\text {precision}}$ and  $\bar{F}_{\text {recall}}$ are the survival functions derived from the distance-based cumulative distribution \cite{sun2023unraveling}.

\paragraph{Implementation Details.} We perform experiments on a 64-bit Linux server with AMD EPYC 7402 24-Core processors and RTX 3090 GPU. The main hyperparameters of our model are listed in Table~\ref{tab:de3}. For all baselines, we adopt the model hyperparameters reported in previous works. The code will be given at https://github.com/xxxxxxx/xxxxxxx.

\begin{table}[h]
\centering
\begin{tabular}{c|c}
\hline
Name         & Choices                          \\ \hline
Window size        & {[}16,32,48,64,80,96,112,128 {]} \\
Feature dimension  & {[}32,64,96,128{]}               \\
Teacher layers     & {[}1,2,3,4,5,6{]}                \\
Student layers     & {[}1,2,3,4,5,6{]}                \\
Prototype number   & {[}16,32,48,64,80{]}             \\
Patch size         & 8                               \\
Head number        & 8                                \\ \hline
Epochs             & 100                              \\
Patience           & 10                  \\
Learning rate      & 0.0001                           \\
Contrastive weight & 0.1                              \\
Batch size         & 128                              \\ \hline
\end{tabular}
\caption{Main hyperparameters}\label{tab:de3}
\end{table}

\begin{table*}[b]
\centering
\scalebox{0.85}{
\setlength{\tabcolsep}{0.25em}
\begin{tabular}{c|llll|llll|llll|llll|llll}
\hline
\multicolumn{1}{l|}{} & \multicolumn{4}{c|}{ABP}                                                                                       & \multicolumn{4}{c|}{Acceleration}                                                                              & \multicolumn{4}{c|}{Air Temperature}                                                                           & \multicolumn{4}{c|}{ECG}                                                                                       & \multicolumn{4}{c}{EPG}                                                                             \\ \cline{2-21} 
\multicolumn{1}{l|}{} & \multicolumn{1}{c}{Acc}   & \multicolumn{1}{c}{AP}    & \multicolumn{1}{c}{AR}    & \multicolumn{1}{c|}{AF1}   & \multicolumn{1}{c}{Acc}   & \multicolumn{1}{c}{AP}    & \multicolumn{1}{c}{AR}    & \multicolumn{1}{c|}{AF1}   & \multicolumn{1}{c}{Acc}   & \multicolumn{1}{c}{AP}    & \multicolumn{1}{c}{AR}    & \multicolumn{1}{c|}{AF1}   & \multicolumn{1}{c}{Acc}   & \multicolumn{1}{c}{AP}    & \multicolumn{1}{c}{AR}    & \multicolumn{1}{c|}{AF1}   & \multicolumn{1}{c}{Acc} & \multicolumn{1}{c}{AP} & \multicolumn{1}{c}{AR} & \multicolumn{1}{c}{AF1} \\ \hline             
Informer     &   0.738                        &     0.810                      &         0.805                  &    0.807                        &    1.000                      &      0.929                     &       0.921                    &        0.925                    &       0.692                    &     0.757                      &           0.752                &    0.754                        &          0.450                 &     0.694                      &     0.693                      &      0.693                      &    0.560                     &      0.830                  &      0.826                  &     0.828                    \\
Reformer &    0.714                       &     0.815                      &    0.807                       &   0.811                         &       0.714                    &       0.732                    &      0.728                     &    0.730                        &     1.000                      &     0.965                      &        0.948                   &      0.956                      &       0.340                    &        0.655                   &        0.654                   &     0.655                       &        0.640                 &      0.882                  &      0.880                  &         0.881                \\
Autoformer &     0.762                      &     0.856                      &     0.851                      &               0.853             &          1.000                &        0.918                   &        0.909                   &     0.913                       &       1.000                    &       0.925                    &       0.913                    &    0.919                       &     0.352                      &     0.613                      &      0.613                     &    0.613                        &    0.680                     &   0.850                     &      0.847                  &    0.848                     \\
TCN &     0.595                      &       0.761                    &      0.750                     &    0.755                        &       1.000                    &     0.929                      &        0.921                   &      0.925                      &     0.692                      &    0.815                       &    0.808                       &        0.812                    &    0.264                       &   0.544                        &     0.544                      &     0.544                       &   0.280                      &     0.680                   &         0.680               &       0.680                  \\
TimesNet     &       0.452                    &        0.664                   &       0.661                   &        0.662                    &    0.571                       &     0.719                      &      0.692                     &         0.705                   &          0.462                 &       0.772                    &      0.777                     &       0.775                     &    0.253                       &   0.562                        &   0.561     &       0.561      &      0.360                     &       0.613                     &   0.613                        &      0.613     \\
TST     &   0.667                        &     0.768                      &        0.756                   &      0.762                      &    0.571                       &     0.733                      &      0.733                     &         0.733                   &          0.923                 &       0.925                    &      0.902                     &       0.913                     &   0.451                        &      0.685                     &  0.683      &      0.684       &      0.440                     &       0.742                     &   0.738                        &      0.740     \\
Nonaug     &     0.595                      &      0.735                     &      0.730                     &     0.732                       &      0.571                     &     0.762                      &     0.761                      &       0.761                     &     0.846                      &      0.778                     &     0.773                      &         0.775                   &             0.230              &      0.575                     &     0.575                      &        0.575                    &                0.360        &     0.652                 &     0.653                   &   0.652                      \\
w/o center     &    0.714                       &      0.868                     &      0.860                     &     0.864       &     0.571                      &      0.746                     &      0.753                     &      0.749                      &       0.846                   &       0.871                   &       0.866                    &        0.868                    &             0.286            &     0.672                      &     0.673                    &      0.672                     &   0.615                    &         0.672           &   0.672                     &       0.672                 \\
w/ feature     &   0.857                        &      0.915                     &      0.904                     &        0.909                    &     0.429                      &     0.684                      &     0.682                     &    0.683                        &         0.769                  &     0.864                      &      0.858                     &      0.861                      &               0.593            &    0.780                      &        0.780                   &    0.780                        &  0.360                       &     0.718                   &    0.717                    &      0.718                   \\
Ours                   &  0.857                     & 0.931                     & 0.910                     & 0.920                      & 1.000                      & 0.965                    & 0.948                   & 0.956                    & 1.000                  & 0.989                     & 0.959                  & 0.974                     & 0.758                   & 0.768                     & 0.808                     & 0.787                      & 0.920                   & 0.935                  & 0.932                  & 0.933                  \\ \hline
                      & \multicolumn{4}{c|}{Gait}                                                                                      & \multicolumn{4}{c|}{NASA}                                                                                      & \multicolumn{4}{c|}{PowerDemand}                                                                               & \multicolumn{4}{c|}{RESP}                                                                                      & \multicolumn{4}{c}{Avg}                                                                             \\ \cline{2-21} 
                      & \multicolumn{1}{c}{Acc}   & \multicolumn{1}{c}{AP}    & \multicolumn{1}{c}{AR}    & \multicolumn{1}{c|}{AF1}   & \multicolumn{1}{c}{Acc}   & \multicolumn{1}{c}{AP}    & \multicolumn{1}{c}{AR}    & \multicolumn{1}{c|}{AF1}   & \multicolumn{1}{c}{Acc}   & \multicolumn{1}{c}{AP}    & \multicolumn{1}{c}{AR}    & \multicolumn{1}{c|}{AF1}   & \multicolumn{1}{c}{Acc}   & \multicolumn{1}{c}{AP}    & \multicolumn{1}{c}{AR}    & \multicolumn{1}{c|}{AF1}   & \multicolumn{1}{c}{Acc} & \multicolumn{1}{c}{AP} & \multicolumn{1}{c}{AR} & \multicolumn{1}{c}{AF1} \\ \hline            
Informer     &    0.485                       &     0.750                      &      0.747                     &     0.748                       &    0.455                       &    0.567                       &      0.567                     &     0.567                       &      0.364                     &         0.561                  &     0.560                      &       0.561                     &        0.353                   &      0.564                     &      0.563                     &     0.563                       &     0.532                    &     0.724                  &      0.720                 &     0.722                   \\
Reformer &     0.636                      &    0.792                       &       0.787                    &            0.790                &       0.818                    &          0.820                 &          0.809                 &     0.815                       &    0.455                       &     0.677                      &     0.675                      &     0.676                       &      0.471                     &    0.599                       &       0.598                   &    0.598                        &     0.552                &   0.745                     &      0.729                  &        0.737                 \\
Autoformer &     0.818                      &    0.881                       &   0.880                        &                 0.880           &         1.000                  &      0.949                     &     0.944                      &   0.946                        &         0.727                  &       0.836                    &      0.833                     &    0.834                        &       0.176                    &      0.597                     &     0.597                      &          0.597                  &     0.600                    &         0.761               &       0.736                 &        0.748                 \\
TCN &    0.667                       &    0.784                       &      0.780                     &              0.782              &      0.909                     &     0.895                      &      0.966                     &    0.929                        &      0.455                     &     0.593                      &     0.592                      &     0.592                       &     0.235                      &       0.559                    &      0.558                     &        0.559                    &          0.452               &    0.669                    &       0.669                &        0.669                 \\
TimesNet     &    0.424                       &       0.682                    &     0.681                      &      0.681                      &    0.273                       &     0.585                      &      0.585                     &         0.585                   &          0.091                 &       0.497                    &      0.495                     &       0.496                     &   0.118                        &      0.393                     &  0.393      &      0.393       &       0.324                  &     0.602                       &    0.600                      &   0.601        \\
TST     &   0.576                        &    0.716                       &      0.710                     &      0.713                      &       0.455                    &     0.596                      &        0.595                   &       0.596                     &             0.455              &    0.623                       &    0.622                       &       0.623                     &              0.176             &        0.557                   &   0.557       &  0.557       &    0.512       &   0.707        &    0.703       &  0.705    \\ 
Nonaug     &    0.636                       &    0.804                       &      0.800                     &      0.802                      &       0.727                    &      0.814                     &       0.812                    &         0.813                  &       0.455                    &    0.700                       &    0.700                       &       0.700                     &            0.353               &          0.748                 &      0.748                    &     0.748                       &         0.440                &          0.683              &        0.670                &         0.676               \\
w/o center     &    0.667                       &     0.803                      &  0.800                         &      0.802                      &    0.545                       &      0.746                     &      0.745                    &    0.745                      &       0.455                    &     0.599                      &    0.596                       &       0.597                     &            0.235              &     0.562                      &     0.561                      &     0.562                       &       0.604                 &           0.727            &      0.710                  &          0.718               \\
w/ feature     &   0.424                       &    0.685                       &     0.682                     &      0.683                    &     0.545                      &      0.738                     &     0.739                      &      0.739                      &       0.545                    &     0.711                     &       0.708                   &     0.709                       &             0.353            &     0.668                      &       0.667                    &       0.667                     &     0.575                  &    0.773                  &       0.771                 &       0.772                  \\
Ours             & 0.878  &  0.891  &    0.852 &  0.871 &   1.000 &  0.969 &   0.953 &  0.961 &  0.818 & 0.888 &  0.884 &    0.886 & 0.471 & 0.736 & 0.736 & 0.736 & 0.820    & 0.857   & 0.860   & 0.858    \\ \hline
\end{tabular}}
\caption{Overall results of ablation studies.}\label{tab:abla}
\vspace{-1.5em}
\end{table*}

\section{Complete Results}
\paragraph{Ablation studies.}
Results of all ablation studies are reported in Table~\ref{tab:abla}. We compare our model with three kinds of variants: (1) Different teacher networks including Informer, Reformer, and Autoformer, which all demonstrate exceptional performance in time series analysis tasks. (2) Different student networks including TCN, TimesNet and TST. (3) Different training strategies including 'Nonaug', 'w/o center', and 'w/feature'. The ablation study validates the effectiveness of the prototype-based student network, the LLM-based teacher network, and the data augmentation-based training strategy. 
\paragraph{Parameter sensitivity.}
We also explore the influence of important hyperparameters on the model performance including window size (Table~\ref{tab:sens1}), data augmentation methods (Table~\ref{tab:sens2}), number of the teacher network (Table~\ref{tab:sens3}), number of the student network (Table~\ref{tab:sens4}) and number of prototypes (Table~\ref{tab:sens5}).

\begin{table*}[h]
\centering
\scalebox{0.85}{
\setlength{\tabcolsep}{0.25em}
\begin{tabular}{c|llll|llll|llll|llll|llll}
\hline
\multicolumn{1}{l|}{} & \multicolumn{4}{c|}{ABP}                                                                                       & \multicolumn{4}{c|}{Acceleration}                                                                              & \multicolumn{4}{c|}{Air Temperature}                                                                           & \multicolumn{4}{c|}{ECG}                                                                                       & \multicolumn{4}{c}{EPG}                                                                             \\ \cline{2-21} 
\multicolumn{1}{l|}{} & \multicolumn{1}{c}{Acc}   & \multicolumn{1}{c}{AP}    & \multicolumn{1}{c}{AR}    & \multicolumn{1}{c|}{AF1}   & \multicolumn{1}{c}{Acc}   & \multicolumn{1}{c}{AP}    & \multicolumn{1}{c}{AR}    & \multicolumn{1}{c|}{AF1}   & \multicolumn{1}{c}{Acc}   & \multicolumn{1}{c}{AP}    & \multicolumn{1}{c}{AR}    & \multicolumn{1}{c|}{AF1}   & \multicolumn{1}{c}{Acc}   & \multicolumn{1}{c}{AP}    & \multicolumn{1}{c}{AR}    & \multicolumn{1}{c|}{AF1}   & \multicolumn{1}{c}{Acc} & \multicolumn{1}{c}{AP} & \multicolumn{1}{c}{AR} & \multicolumn{1}{c}{AF1} \\ \hline             
16     &   0.738                        &     0.871                      &         0.867                  &    0.869                        &    1.000                      &      0.958                     &       0.934                    &        0.946                    &       1.000                    &     0.912                      &           0.905                &    0.909                        &          0.769                 &     0.751                      &     0.750                      &      0.750                      &    0.720                     &      0.795                  &      0.791                  &     0.793                    \\
32                &  0.857                     & 0.931                     & 0.910                     & 0.920                      & 1.000                      & 0.965                    & 0.948                   & 0.956                    & 1.000                  & 0.989                     & 0.959                  & 0.974                     & 0.758                   & 0.768                     & 0.808                     & 0.787                      & 0.920                   & 0.935                  & 0.932                  & 0.933                               \\
48   &     0.785                      &     0.863                      &     0.858                      &               0.861             &          1.000                &        0.956                   &        0.950                   &     0.953                       &       1.000                    &       0.965                    &       0.957                    &    0.961                       &     0.747                      &     0.818                      &      0.817                     &    0.818                        &    0.840                     &   0.904                     &      0.904                  &    0.904                     \\
64 &      0.785        &  0.867           &    0.862    &   0.865    &   1.000      &  0.918    &  0.909    &  0.913    &  1.000    &  0.925    &  0.913    &  0.919    &  0.769    &  0.780    &  0.779    &  0.780
    &  0.720    &  0.860    &  0.858    &  0.859  \\
80    &       0.857      &     0.935  &     0.931  &     0.933   &     1.000  &     0,970  &     0.948  &     0.959   &     1.000  &     0.935  &     0.924  &     0.929  &     0.758  &     0.783  &     0.782  &     0.782   &     0.880  &     0.906  &     0.904  &     0.905 \\
96         &      0.904    &      0.954    &      0.846    &      0.950     &      1.000    &      0.898    &      0.900    &      0.899
    &      1.000    &      0.904    &      0.901    &      0.903 &      0.791    &      0.851    &      0.848    &      0.850 &      0.960   &      0.960   &      0.959   &      0.960    \\
112        &    0.833   &    0.946   &    0.937   &    0.941  &    1.000   &    0.920   &    0.910   &    0.905
   &    1.000   &    0.930   &    0.924   &    0.927  &    0.758   &    0.772   &    0.771   &    0.771  &    0.880   &    0.927   &    0.923   &    0.925  \\
128          &    0.833     &    0.903     &    0.895     &    0.899    &    1.000     &    0.919     &    0.910     &    0.914
     &    1.000     &    0.935     &    0.923     &    0.929   &    0.780     &    0.787     &    0.785     &    0.786  &    0.720     &    0.898     &    0.896     &    0.897   
\\ \hline
                      & \multicolumn{4}{c|}{Gait}                                                                                      & \multicolumn{4}{c|}{NASA}                                                                                      & \multicolumn{4}{c|}{PowerDemand}                                                                               & \multicolumn{4}{c|}{RESP}                                                                                      & \multicolumn{4}{c}{Avg}                                                                             \\ \cline{2-21} 
                      & \multicolumn{1}{c}{Acc}   & \multicolumn{1}{c}{AP}    & \multicolumn{1}{c}{AR}    & \multicolumn{1}{c|}{AF1}   & \multicolumn{1}{c}{Acc}   & \multicolumn{1}{c}{AP}    & \multicolumn{1}{c}{AR}    & \multicolumn{1}{c|}{AF1}   & \multicolumn{1}{c}{Acc}   & \multicolumn{1}{c}{AP}    & \multicolumn{1}{c}{AR}    & \multicolumn{1}{c|}{AF1}   & \multicolumn{1}{c}{Acc}   & \multicolumn{1}{c}{AP}    & \multicolumn{1}{c}{AR}    & \multicolumn{1}{c|}{AF1}   & \multicolumn{1}{c}{Acc} & \multicolumn{1}{c}{AP} & \multicolumn{1}{c}{AR} & \multicolumn{1}{c}{AF1} \\ \hline            
16     &    0.788                       &     0.867                      &      0.864                     &     0.866                       &    1.000                       &    0.958                       &      0.952                     &     0.955                       &      0.818                     &         0.870                  &     0.866                      &       0.868                     &        0.176                   &      0.502                     &      0.502                     &     0.502                       &       0.752                  &            0.803            &     0.800                   &       0.801                  \\
32      & 0.878  &  0.891  &    0.852 &  0.871 &   1.000 &  0.969 &   0.953 &  0.961 &  0.818 & 0.888 &  0.884 &    0.886 & 0.471 & 0.736 & 0.736 & 0.736 & 0.820    & 0.857   & 0.860   & 0.858        \\
48  &     0.848                      &    0.775                       &   0.772                        &                 0.773           &        1.000                  &      0.949                     &     0.946                      &    0.947                        &         0.727                  &       0.882                    &      0.919                     &    0.900                        &       0.294                    &      0.738                     &     0.738                      &          0.738                  &     0.780                    &               0.843         &        0.842               &          0.842               \\
64  &      0.727         &        0.807          &       0.805           &     0.806       &    1.000        &       0.949    &      0.945    &    0.946     &     0.636    &    0.873    &    0.870    &    0.871          &    0.471    &    0.774    &    0.774    &    0.774  &  0.762        & 0.837  &  0.835    &0.836  \\
80           &      0.848      &      0.785      &      0.781      &      0.783 &      0.727      &      0.947      &      0.944      &      0.945
&      0.727      &      0.849      &      0.845      &      0.847  &      0.353      &      0.774      &      0.774      &      0.774     
&   0.784    &  0.844  &  0.841   &  0.842     \\
96             &       0.848        &       0.889        &       0.884        &       0.885  &       0.909        &       0.945        &       0.941        &       0.943&       0.818        &       0.902        &       0.899        &       0.900    &       0.353        &       0.778        &       0.778        &       0.778    &    0.828     &   0.890        &   0.886        &  0.888   \\ 
112         &     0.878    &     0.859    &     0.858    &     0.858  &     0.909    &     0.971    &     0.960    &     0.966 &     0.818    &     0.880    &     0.878    &     0.879   &     0.353    &     0.753    &     0.743    &     0.743 & 0.768    & 0.849  &  0.844  &  0.846 \\
128             &      0.758      &     0.873      &     0.885      &     0.878  &     1.000      &     0.959      &     0.951      &     0.955
      &     0.727      &     0.837      &     0.835      &     0.836    &     0.235      &     0.760      &     0.760      &     0.760   & 0.716 & 0.827  &  0.826   &  0.826
    \\ \hline
\end{tabular}}
\caption{Overall results of different window sizes.}\label{tab:sens1}
\vspace{-1.5em}
\end{table*}

\begin{table*}[t]
\centering
\scalebox{0.85}{
\setlength{\tabcolsep}{0.25em}
\begin{tabular}{c|llll|llll|llll|llll|llll}
\hline
\multicolumn{1}{l|}{} & \multicolumn{4}{c|}{ABP}                                                                                       & \multicolumn{4}{c|}{Acceleration}                                                                              & \multicolumn{4}{c|}{Air Temperature}                                                                           & \multicolumn{4}{c|}{ECG}                                                                                       & \multicolumn{4}{c}{EPG}                                                                             \\ \cline{2-21} 
\multicolumn{1}{l|}{} & \multicolumn{1}{c}{Acc}   & \multicolumn{1}{c}{AP}    & \multicolumn{1}{c}{AR}    & \multicolumn{1}{c|}{AF1}   & \multicolumn{1}{c}{Acc}   & \multicolumn{1}{c}{AP}    & \multicolumn{1}{c}{AR}    & \multicolumn{1}{c|}{AF1}   & \multicolumn{1}{c}{Acc}   & \multicolumn{1}{c}{AP}    & \multicolumn{1}{c}{AR}    & \multicolumn{1}{c|}{AF1}   & \multicolumn{1}{c}{Acc}   & \multicolumn{1}{c}{AP}    & \multicolumn{1}{c}{AR}    & \multicolumn{1}{c|}{AF1}   & \multicolumn{1}{c}{Acc} & \multicolumn{1}{c}{AP} & \multicolumn{1}{c}{AR} & \multicolumn{1}{c}{AF1} \\ \hline             
J          &      0.738      &     0.797      &     0.792      &     0.794   &     1.000      &     0.909      &     0.903      &     0.906
      &     1.000      &     0.892      &     0.887      &     0.889    &     0.538      &     0.738      &     0.736      &     0.737
      &     0.800      &     0.881      &     0.880      &     0.880  \\
S       &     0.761     &     0.849     &     0.843     &     0.846   &     0.857     &     0.845     &     0.826     &     0.835  &     1.000     &     0.923     &     0.913     &     0.918  &     0.637     &     0.793     &     0.791     &     0.792 &     0.880     &     0.920     &     0.919     &     0.919    \\
W      &     0.738     &     0.813     &     0.810     &     0.811   &     1.000     &     0.917     &     0.909     &     0.913  &     1.000     &     0.896     &     0.896     &     0.896   &     0.549     &     0.764     &     0.764     &     0.764  &     0.800     &     0.881     &     0.877     &     0.879   \\
J + S       &     0.810    &     0.821    &     0.812    &     0.816  &     1.000    &     0.944    &     0.925    &     0.935 &     1.000    &     0.894    &     0.892    &     0.893  &     0.495    &     0.738    &     0.737    &     0.738 &     0.800    &     0.903    &     0.902    &     0.903\\
J + W   &    0.857    &    0.931    &    0.910    &    0.920 &    1.000    &    0.965    &    0.948    &    9.956 &    1.000    &    0.989    &    0.959    &    0.974  &    0.758    &    0.768    &    0.808    &    0.787  &    0.920    &    0.935    &    0.932    &    0.933        \\
S + W         &    0.810    &    0.906    &    0.903    &    0.905 &    1.000    &    0.945    &    0.940    &    0.942 &    1.000    &    0.901    &    0.896    &    0.899  &    0.769    &    0.798    &    0.796    &    0.797  &    0.920    &    0.939    &    0.936    &    0.938 \\
J + S + W          &    0.857    &    0.918    &    0.910    &    0.914  &    1.000    &    0.944    &    0.941    &    0.943 &    1.000    &    0.914    &    0.906    &    0.910  &    0.769    &    0.739    &    0.738    &    0.738 &    0.800    &    0.922    &    0.919    &    0.920
              \\ \hline
                      & \multicolumn{4}{c|}{Gait}                                                                                      & \multicolumn{4}{c|}{NASA}                                                                                      & \multicolumn{4}{c|}{PowerDemand}                                                                               & \multicolumn{4}{c|}{RESP}                                                                                      & \multicolumn{4}{c}{Avg}                                                                             \\ \cline{2-21} 
                      & \multicolumn{1}{c}{Acc}   & \multicolumn{1}{c}{AP}    & \multicolumn{1}{c}{AR}    & \multicolumn{1}{c|}{AF1}   & \multicolumn{1}{c}{Acc}   & \multicolumn{1}{c}{AP}    & \multicolumn{1}{c}{AR}    & \multicolumn{1}{c|}{AF1}   & \multicolumn{1}{c}{Acc}   & \multicolumn{1}{c}{AP}    & \multicolumn{1}{c}{AR}    & \multicolumn{1}{c|}{AF1}   & \multicolumn{1}{c}{Acc}   & \multicolumn{1}{c}{AP}    & \multicolumn{1}{c}{AR}    & \multicolumn{1}{c|}{AF1}   & \multicolumn{1}{c}{Acc} & \multicolumn{1}{c}{AP} & \multicolumn{1}{c}{AR} & \multicolumn{1}{c}{AF1} \\ \hline            
J         &     0.667    &     0.759    &     0.756    &     0.758  &     1.000    &     0.954    &     0.949    &     0.951  &     0.636    &     0.870    &     0.865    &     0.868   &     0.412    &     0.684    &     0.683    &     0.684   
&  0.668  &  0.790  &  0.787   & 0.788 \\
S      &    0.636     &    0.777     &    0.773     &    0.775   &    0.909     &    0.920     &    0.916     &    0.918   &    0.727     &    0.829     &    0.826     &    0.828   &    0.294     &    0.540     &    0.540     &    0.540 
& 0.700   &  0.804  &  0.813   &  0.809  \\
W         &    0.636    &    0.774    &    0.770    &    0.772  &    1.000    &    0.967    &    0.960    &    0.964
    &    0.909    &    0.957    &    0.949    &    0.953   &    0.529    &    0.709    &    0.709    &    0.709  
    &  0.809  &  0.805   &   0.813  &   0.809    \\
J + S     &    0.667     &    0.796     &    0.794     &    0.795  &    1.000     &    0.942     &    0.937     &    0.940  &    0.727     &    0.760     &    0.757     &    0.759   &    0.471     &    0.761     &    0.760     &    0.761
& 0.672   &  0.801  &  0.800   &  0.800   \\
J + W
&    0.878  &    0.861  &    0.852  &    0.857   &    1.000  &    0.969  &    0.953  &    0.961   &    0.818  &    0.888  &    0.884  &    0.886    &    0.471  &    0.736  &    0.736  &    0.736
&  0.820  &   0.857 &  0.860   & 0.858   \\
S + W         &    0.878    &    0.871    &    0.870    &    0.871   &    1.000    &    0.972    &    0.971    &    0.971
    &    0.818    &    0.873    &    0.872    &    0.873   &    0.471    &    0.764    &    0.763    &    0.763  
    &  0.816  &   0.867 & 0.863    &  0.865   \\ 
J + S + W            &   0.848    &   0.893    &   0.889    &   0.891  &   1.000    &   0.958    &   0.956    &   0.957  &   0.909    &   0.816    &   0.810    &   0.813   &   0.529    &   0.727    &   0.726    &   0.727
& 0.816   & 0.830   &  0.843   &  0.837
    \\ \hline
\end{tabular}}
\caption{Overall results of different augmentation methods.}\label{tab:sens2}
\vspace{-1.5em}
\end{table*}

\begin{table*}[t]
\centering
\scalebox{0.85}{
\setlength{\tabcolsep}{0.25em}
\begin{tabular}{c|llll|llll|llll|llll|llll}
\hline
\multicolumn{1}{l|}{} & \multicolumn{4}{c|}{ABP}                                                                                       & \multicolumn{4}{c|}{Acceleration}                                                                              & \multicolumn{4}{c|}{Air Temperature}                                                                           & \multicolumn{4}{c|}{ECG}                                                                                       & \multicolumn{4}{c}{EPG}                                                                             \\ \cline{2-21} 
\multicolumn{1}{l|}{} & \multicolumn{1}{c}{Acc}   & \multicolumn{1}{c}{AP}    & \multicolumn{1}{c}{AR}    & \multicolumn{1}{c|}{AF1}   & \multicolumn{1}{c}{Acc}   & \multicolumn{1}{c}{AP}    & \multicolumn{1}{c}{AR}    & \multicolumn{1}{c|}{AF1}   & \multicolumn{1}{c}{Acc}   & \multicolumn{1}{c}{AP}    & \multicolumn{1}{c}{AR}    & \multicolumn{1}{c|}{AF1}   & \multicolumn{1}{c}{Acc}   & \multicolumn{1}{c}{AP}    & \multicolumn{1}{c}{AR}    & \multicolumn{1}{c|}{AF1}   & \multicolumn{1}{c}{Acc} & \multicolumn{1}{c}{AP} & \multicolumn{1}{c}{AR} & \multicolumn{1}{c}{AF1} \\ \hline             
1    &    0.714      &    0.745      &    0.738      &    0.742 &    1.000      &    0.900      &    0.897      &    0.898  &    1.000      &    0.925      &    0.914      &    0.920 &    0.451      &    0.709      &    0.708      &    0.709  &    0.720      &    0.774      &    0.770      &    0.772             \\
2         &    0.857           &    0.881           &    0.872           &    0.876 &    1.000           &    0.970           &    0.956           &    0.963  &    1.000           &    0.917           &    0.911           &    0.914 &    0.538           &    0.723           &    0.721           &    0.722  &    0.840           &    0.882           &    0.881           &    0.882\\
3      &    0.857     &    0.931     &    0.910     &    0.920 &    1.000     &    0.965     &    0.948       &    0.956 &   1.000    &   0.989    &   0.959    &   0.974 &   0.758    &   0.768    &   0.808    &   0.787  &   0.920    &   0.935    &   0.932    &   0.933  \\
4     &    0.880    &    0.909    &    0.898    &    0.904 
&    1.000    &    0.937    &    0.924    &    0.930 
&    1.000    &    0.917    &    0.904    &    0.911  
&    0.769    &    0.785    &    0.783    &    0.784 
&    0.880    &    0.924    &    0.921    &    0.923  \\
5       &    0.833  &    0.904  &    0.899  &    0.901 
&    1.000  &    0.948  &    0.931  &    0.940 
&    1.000  &    0.906  &    0.901  &    0.903  
&    0.769  &    0.760  &    0.758  &    0.759 
&    0.920  &    0.921  &    0.919  &    0.920  \\
6          &    0.880    &    0.876    &    0.865    &    0.870 
&    1.000    &    0.878    &    0.883    &    0.881 
&    1.000    &    0.928    &    0.917    &    0.923 
&    0.791    &    0.815    &    0.813    &    0.814 
&    0.960    &    0.909    &    0.908    &    0.908
                   \\ \hline
                      & \multicolumn{4}{c|}{Gait}                                                                                      & \multicolumn{4}{c|}{NASA}                                                                                      & \multicolumn{4}{c|}{PowerDemand}                                                                               & \multicolumn{4}{c|}{RESP}                                                                                      & \multicolumn{4}{c}{Avg}                                                                             \\ \cline{2-21} 
                      & \multicolumn{1}{c}{Acc}   & \multicolumn{1}{c}{AP}    & \multicolumn{1}{c}{AR}    & \multicolumn{1}{c|}{AF1}   & \multicolumn{1}{c}{Acc}   & \multicolumn{1}{c}{AP}    & \multicolumn{1}{c}{AR}    & \multicolumn{1}{c|}{AF1}   & \multicolumn{1}{c}{Acc}   & \multicolumn{1}{c}{AP}    & \multicolumn{1}{c}{AR}    & \multicolumn{1}{c|}{AF1}   & \multicolumn{1}{c}{Acc}   & \multicolumn{1}{c}{AP}    & \multicolumn{1}{c}{AR}    & \multicolumn{1}{c|}{AF1}   & \multicolumn{1}{c}{Acc} & \multicolumn{1}{c}{AP} & \multicolumn{1}{c}{AR} & \multicolumn{1}{c}{AF1} \\ \hline            
1    &   0.667      &   0.808      &   0.806      &   0.807
      &   0.909      &   0.864      &   0.861      &   0.862
      &   0.818      &   0.846      &   0.844      &   0.845
      &   0.235      &   0.622      &   0.622      &   0.622     
&  0.626   &   0.756    & 0.758       &  0.757      \\
2       
      &   0.758      &   0.886      &   0.881      &   0.884  
&   0.909      &   0.935      &   0.923      &   0.929 
&   0.818      &   0.919      &   0.916      &   0.918   
&   0.471      &   0.754      &   0.753      &   0.753
    & 0.712    &   0.820    &   0.824     & 0.822
   \\
3      &   0.878     &   0.861     &   0.852     &   0.857
     &   1.000     &   0.969     &   0.953     &   0.961
     &   0.818     &   0.888     &   0.884     &   0.886
     &   0.471     &   0.736     &   0.736     &   0.736
     &  0.820   &  0.857     &   0.860     & 0.858
                \\
4      &   0.848     &   0.863     &   0.861     &   0.862
     &   1.000     &   0.966     &   0.956     &   0.961
     &   0.818     &   0.936     &   0.934     &   0.935
     &   0.412     &   0.731     &   0.731     &   0.731
     & 0.816    &   0.851    &   0.850     & 0.850
                 \\
5        &   0.848    &   0.821    &   0.849    &   0.850
    &   1.000    &   0.940    &   0.938    &   0.939
    &   0.818    &   0.918    &   0.916    &   0.917
    &   0.412    &   0.732    &   0.732    &   0.732
    &  0.812   &  0.836     &   0.839     & 0.837
     \\
6                     &    0.788     &   0.818     &   0.817     &   0.818
     &   1.000     &   0.949     &   0.943     &   0.946
     &   0.727     &   0.888     &   0.883     &   0.885
     &   0.412     &   0.823     &   0.823     &   0.823
     & 0.820    &   0.851    &   0.849     &   0.850
    \\ \hline
\end{tabular}}
\caption{Overall results of different different teacher layers.}\label{tab:sens3}
\vspace{-1.5em}
\end{table*}

\begin{table*}[t]
\centering
\scalebox{0.85}{
\setlength{\tabcolsep}{0.25em}
\begin{tabular}{c|llll|llll|llll|llll|llll}
\hline
\multicolumn{1}{l|}{} & \multicolumn{4}{c|}{ABP}                                                                                       & \multicolumn{4}{c|}{Acceleration}                                                                              & \multicolumn{4}{c|}{Air Temperature}                                                                           & \multicolumn{4}{c|}{ECG}                                                                                       & \multicolumn{4}{c}{EPG}                                                                             \\ \cline{2-21} 
\multicolumn{1}{l|}{} & \multicolumn{1}{c}{Acc}   & \multicolumn{1}{c}{AP}    & \multicolumn{1}{c}{AR}    & \multicolumn{1}{c|}{AF1}   & \multicolumn{1}{c}{Acc}   & \multicolumn{1}{c}{AP}    & \multicolumn{1}{c}{AR}    & \multicolumn{1}{c|}{AF1}   & \multicolumn{1}{c}{Acc}   & \multicolumn{1}{c}{AP}    & \multicolumn{1}{c}{AR}    & \multicolumn{1}{c|}{AF1}   & \multicolumn{1}{c}{Acc}   & \multicolumn{1}{c}{AP}    & \multicolumn{1}{c}{AR}    & \multicolumn{1}{c|}{AF1}   & \multicolumn{1}{c}{Acc} & \multicolumn{1}{c}{AP} & \multicolumn{1}{c}{AR} & \multicolumn{1}{c}{AF1} \\ \hline             
1           &     0.690      &     0.803      &     0.799      &     0.801
      &     1.000      &     0.906      &     0.909      &     0.907
      &     1.000      &     0.908      &     0.903      &     0.906
      &     0.527      &     0.707      &     0.705      &     0.706
      &     0.880      &     0.916      &     0.913      &     0.914
                    \\
2      &    0.714    &    0.788    &    0.785    &    0.787
    &    1.000    &    0.915    &    0.909    &    0.912
    &    1.000    &    0.873    &    0.878    &    0.875
    &    0.516    &    0.741    &    0.740    &    0.740
    &    0.720    &    0.848    &    0.847    &    0.847
              \\
3      &    0.857     &    0.931     &    0.910     &    0.920 &    1.000     &    0.965     &    0.948       &    0.956 &   1.000    &   0.989    &   0.959    &   0.974 &   0.758    &   0.768    &   0.808    &   0.787  &   0.920    &   0.935    &   0.932    &   0.933  \\
4       &     0.714      &     0.770      &     0.767      &     0.768
      &     0.857      &     0.907      &     0.897      &     0.902
      &     1.000      &     0.866      &     0.870      &     0.868
      &     0.505      &     0.730      &     0.730      &     0.730
      &     0.640      &     0.802      &     0.804      &     0.803
                  \\
5    &     0.738      &     0.800      &     0.797      &     0.799
      &     0.857      &     0.925      &     0.902      &     0.914
      &     1.000      &     0.882      &     0.881      &     0.881
      &     0.538      &     0.736      &     0.735      &     0.736
      &     0.640      &     0.724      &     0.722      &     0.723    
    \\
6          &    0.714    &    0.765    &    0.762    &    0.763
    &    1.000    &    0.886    &    0.882    &    0.884
    &    1.000    &    0.892    &    0.888    &    0.890
    &    0.451    &    0.702    &    0.702    &    0.702
    &    0.640    &    0.749    &    0.748    &    0.748
                   \\ \hline
                      & \multicolumn{4}{c|}{Gait}                                                                                      & \multicolumn{4}{c|}{NASA}                                                                                      & \multicolumn{4}{c|}{PowerDemand}                                                                               & \multicolumn{4}{c|}{RESP}                                                                                      & \multicolumn{4}{c}{Avg}                                                                             \\ \cline{2-21} 
                      & \multicolumn{1}{c}{Acc}   & \multicolumn{1}{c}{AP}    & \multicolumn{1}{c}{AR}    & \multicolumn{1}{c|}{AF1}   & \multicolumn{1}{c}{Acc}   & \multicolumn{1}{c}{AP}    & \multicolumn{1}{c}{AR}    & \multicolumn{1}{c|}{AF1}   & \multicolumn{1}{c}{Acc}   & \multicolumn{1}{c}{AP}    & \multicolumn{1}{c}{AR}    & \multicolumn{1}{c|}{AF1}   & \multicolumn{1}{c}{Acc}   & \multicolumn{1}{c}{AP}    & \multicolumn{1}{c}{AR}    & \multicolumn{1}{c|}{AF1}   & \multicolumn{1}{c}{Acc} & \multicolumn{1}{c}{AP} & \multicolumn{1}{c}{AR} & \multicolumn{1}{c}{AF1} \\ \hline            
1          &     0.697      &    0.816      &    0.813      &    0.814
      &    1.000      &    0.951      &    0.949      &    0.950
      &    0.818      &    0.853      &    0.848      &    0.851
      &    0.294      &    0.682      &    0.682      &    0.682
      & 0.668   &  0.791   &   0.788   &0.789
                   \\
2      &    0.758     &    0.876     &    0.882     &    0.879
     &    1.000     &    0.939     &    0.936     &    0.937
     &    0.727     &    0.871     &    0.869     &    0.870
     &    0.471     &    0.682     &    0.681     &    0.682
           &  0.668  &  0.800   &   0.798   & 0.799
                     \\
3       &    0.878      &    0.861      &    0.852      &    0.857
      &    1.000      &    0.969      &    0.953      &    0.961
      &    0.818      &    0.888      &    0.884      &    0.886
      &    0.471      &    0.736      &    0.736      &    0.736
            & 0.820   &  0.857   & 0.860     & 0.858
                 \\
4      &    0.758     &    0.881     &    0.877     &    0.879
     &    0.909     &    0.857     &    0.858     &    0.857
     &    0.636     &    0.711     &    0.711     &    0.711
     &    0.235     &    0.581     &    0.580     &    0.580
           & 0.640   &  0.772   &  0.768    & 0.770
                 \\
5          &     0.606     &    0.720     &    0.720     &    0.720
     &    1.000     &    0.942     &    0.939     &    0.941
     &    0.636     &    0.815     &    0.813     &    0.814
     &    0.412     &    0.768     &    0.768     &    0.768 
           & 0.628   &  0.769   &   0.771   & 0.770
       \\
6               &    0.697     &    0.811     &    0.813     &    0.812
     &    0.909     &    0.859     &    0.858     &    0.859
     &    0.818     &    0.887     &    0.885     &    0.886
     &    0.471     &    0.698     &    0.697     &    0.698
           &  0.628  &  0.770   &  0.760    & 0.761
    \\ \hline
\end{tabular}}
\caption{Overall results of different student layers.}\label{tab:sens4}
\vspace{-1.5em}
\end{table*}

\begin{table*}[t]
\centering
\scalebox{0.85}{
\setlength{\tabcolsep}{0.25em}
\begin{tabular}{c|llll|llll|llll|llll|llll}
\hline
\multicolumn{1}{l|}{} & \multicolumn{4}{c|}{ABP}                                                                                       & \multicolumn{4}{c|}{Acceleration}                                                                              & \multicolumn{4}{c|}{Air Temperature}                                                                           & \multicolumn{4}{c|}{ECG}                                                                                       & \multicolumn{4}{c}{EPG}                                                                             \\ \cline{2-21} 
\multicolumn{1}{l|}{} & \multicolumn{1}{c}{Acc}   & \multicolumn{1}{c}{AP}    & \multicolumn{1}{c}{AR}    & \multicolumn{1}{c|}{AF1}   & \multicolumn{1}{c}{Acc}   & \multicolumn{1}{c}{AP}    & \multicolumn{1}{c}{AR}    & \multicolumn{1}{c|}{AF1}   & \multicolumn{1}{c}{Acc}   & \multicolumn{1}{c}{AP}    & \multicolumn{1}{c}{AR}    & \multicolumn{1}{c|}{AF1}   & \multicolumn{1}{c}{Acc}   & \multicolumn{1}{c}{AP}    & \multicolumn{1}{c}{AR}    & \multicolumn{1}{c|}{AF1}   & \multicolumn{1}{c}{Acc} & \multicolumn{1}{c}{AP} & \multicolumn{1}{c}{AR} & \multicolumn{1}{c}{AF1} \\ \hline             
16           &      0.714      &      0.765      &      0.763      &      0.764
      &      1.000      &      0.872      &      0.873      &      0.873
      &      1.000      &      0.892      &      0.893      &      0.892
      &      0.451      &      0.704      &      0.703      &      0.704
      &      0.600      &      0.780      &      0.780      &      0.780
                   \\
32        &      0.857      &      0.931      &      0.910      &      0.920
      &      1.000      &      0.965      &      0.948      &      9.956
      &      1.000      &      0.989      &      0.959      &      0.974
      &      0.758      &      0.768      &      0.808      &      0.787
      &      0.920      &      0.935      &      0.932      &      0.933
                \\
48      &      0.880     &      0.964     &      0.956     &      0.960
     &      1.000     &      0.946     &      0.932     &      0.939
     &      1.000     &      0.987     &      0.983     &      0.985
     &      0.769     &      0.793     &      0.792     &      0.792
     &      0.920     &      0.948     &      0.945     &      0.947
                     \\
64        &  0.857       &  0.942       &  0.948       &  0.940
       &  1.000       &  0.953       &  0.953       &  0.953
       &  1.000       &  0.977       &  0.976       &  0.977
       &  0.769       &  0.759       &  0.758       &  0.759
       &  0.880       &  0.936       &  0.937       &  0.937
                  \\
80              &             0.809     &     0.853     &     0.849     &     0.851
     &     1.000     &     0.943     &     0.929     &     0.936
     &     1.000     &     0.922     &     0.907     &     0.914
     &     0.758     &     0.715     &     0.714     &     0.714
     &     0.920     &     0.796     &     0.794     &     0.795
                   \\ \hline
                      & \multicolumn{4}{c|}{Gait}                                                                                      & \multicolumn{4}{c|}{NASA}                                                                                      & \multicolumn{4}{c|}{PowerDemand}                                                                               & \multicolumn{4}{c|}{RESP}                                                                                      & \multicolumn{4}{c}{Avg}                                                                             \\ \cline{2-21} 
                      & \multicolumn{1}{c}{Acc}   & \multicolumn{1}{c}{AP}    & \multicolumn{1}{c}{AR}    & \multicolumn{1}{c|}{AF1}   & \multicolumn{1}{c}{Acc}   & \multicolumn{1}{c}{AP}    & \multicolumn{1}{c}{AR}    & \multicolumn{1}{c|}{AF1}   & \multicolumn{1}{c}{Acc}   & \multicolumn{1}{c}{AP}    & \multicolumn{1}{c}{AR}    & \multicolumn{1}{c|}{AF1}   & \multicolumn{1}{c}{Acc}   & \multicolumn{1}{c}{AP}    & \multicolumn{1}{c}{AR}    & \multicolumn{1}{c|}{AF1}   & \multicolumn{1}{c}{Acc} & \multicolumn{1}{c}{AP} & \multicolumn{1}{c}{AR} & \multicolumn{1}{c}{AF1} \\ \hline            
16          &     0.697     &     0.836     &     0.835     &     0.836
     &     1.000     &     0.944     &     0.944     &     0.944
     &     0.636     &     0.810     &     0.808     &     0.809
     &     0.294     &     0.679     &     0.679     &     0.679
     &   0.608    &  0.765      &  0.769       & 0.767
                   \\
32         &    0.878        &    0.861        &    0.852        &    0.857
   &    1.000   &    0.969   &    0.953   &    0.961
   &    0.818   &    0.888   &    0.884   &    0.886
   &    0.471   &    0.736   &    0.736   &    0.736
        &   0.820    &   0.857     &   0.860      &    0.858
                     \\
48     &    0.848    &    0.865    &    0.873    &    0.869
    &    1.000    &    0.935    &    0.934    &    0.934
    &    0.909    &    0.973    &    0.967    &    0.970
    &    0.294    &    0.800    &    0.800    &    0.800
         &  0.820     &  0.876      &  0.875       & 0.875
                 \\
64     &    0.810    &    0.884    &    0.882    &    0.883
    &    1.000    &    0.949    &    0.947    &    0.948
    &    0.909    &    0.968    &    0.965    &    0.967
    &    0.353    &    0.766    &    0.766    &    0.766
         &   0.808    &   0.850     &    0.846     & 0.848
                 \\
80              &    0.878    &    0.881    &    0.876    &    0.879
    &    1.000    &    0.956    &    0.947    &    0.951
    &    0.909    &    0.725    &    0.725    &    0.725
    &    0.471    &    0.680    &    0.679    &    0.680
         &   0.816    &   0.792     &   0.792      & 0.792
    \\ \hline
\end{tabular}}
\caption{Overall results of different prototype sizes.}\label{tab:sens5}
\vspace{-1.5em}
\end{table*}

\bibliographystyle{named}
\bibliography{ijcai23}

\end{document}